\documentclass[reprint, cha]{revtex4-1}

\usepackage{amssymb}
\usepackage{graphicx}
\usepackage{amsmath}
\usepackage{amsfonts}
\usepackage{amssymb}
\usepackage{amsthm}
\usepackage{verbatim}
\usepackage{enumerate}
\usepackage{natbib}
\usepackage{mdwlist} % for compact lists
\usepackage{subfigure}
\usepackage{color}
\usepackage[utf8]{inputenc}
\newcommand{\npar}{\par \vspace{2.3ex plus 0.3ex minus 0.3ex}}
\usepackage{algpseudocode}
\usepackage{algorithmicx}
\usepackage{lineno}

\begin{document}

\title{Improved texture image classification through the use of a corrosion-inspired  cellular automaton}

\author{N\'ubia Rosa da Silva}
 	     \email{nubiasrosa@gmail.com}
\affiliation{Institute of Mathematics and Computer Science, University of S\~{a}o Paulo (USP), Avenida Trabalhador s\~{a}o-carlense, 400 13566-590 S\~{a}o Carlos, S\~{a}o Paulo, Brazil} 
\affiliation{Scientific Computing Group, S\~ao Carlos Institute of Physics, University of S\~{a}o Paulo (USP),  cx 369 13560-970 S\~{a}o Carlos, S\~{a}o Paulo, Brazil - www.scg.ifsc.usp.br}

\author{Pieter Van der Wee\"{e}n}
 	     \email{pieter.vanderween@ugent.be}
\affiliation{Department of Mathematical Modeling, Statistics and Bioinformatics, Ghent University, Coupure links 653, 9000 Ghent, Belgium} 

\author{Bernard De Baets}
 	     \email{bernard.debaets@ugent.be}
\affiliation{Department of Mathematical Modeling, Statistics and Bioinformatics, Ghent University, Coupure links 653, 9000 Ghent, Belgium} 

\author{Odemir M. Bruno}
              \email{bruno@ifsc.usp.br}
\affiliation{Scientific Computing Group, S\~ao Carlos Institute of Physics, University of S\~{a}o Paulo (USP),  cx 369 13560-970 S\~{a}o Carlos, S\~{a}o Paulo, Brazil - www.scg.ifsc.usp.br}

\date{\today}

\begin{abstract}
In this paper, the problem of classifying synthetic and natural texture images is addressed. To tackle this problem, an innovative method is proposed that combines concepts from corrosion modeling and cellular automata to generate a texture descriptor. The core processes of metal (pitting) corrosion are identified and applied to texture images by incorporating the basic mechanisms of corrosion in the transition function of the cellular automaton. The surface morphology of the image is analyzed before and during the application of the transition function of the cellular automaton. In each iteration the cumulative mass of corroded product is obtained to construct each of the attributes of the texture descriptor. In a final step, this texture descriptor is used for image classification by applying Linear Discriminant Analysis. The method was tested on the well-known Brodatz and Vistex databases. In addition, in order to verify the robustness of the method, its invariance to noise and rotation were tested. To that end, different variants of the original two databases were obtained through addition of noise to and rotation of the images. The results showed that the method is effective for texture classification according to the high success rates obtained in all cases. This indicates the potential of employing methods inspired on natural phenomena in other fields.
\end{abstract}

\keywords{
Pattern Recognition, pitting corrosion, texture classification, cellular automata
}

\maketitle

\section{Introduction} 
\label{sec:intro}

The classification of texture images is an important problem in pattern recognition and consequently forms the subject of many research works in this field. Texture is an important image feature with a strong discriminative capability and is therefore widely used in computer vision. The texture classification problem addressed in this paper is a multiclass classification problem and two different well-known databases are considered: the Brodatz database, which contains a number of unique textures that each form a class, and the Vistex database, which consists of a number of classes, each with several texture images belonging to it. For the Brodatz database, ten subimages of the same size are taken from each texture to obtain different texture images for the corresponding class. For each texture image of both databases a feature vector, i.e.\ a vector of characteristics, is obtained by using a novel method introduced in this paper, as well as by using a number of popular methods from literature. This feature vector is then employed to classify the different texture images using Linear Discriminant Analysis (LDA) following a stratified 10-fold cross-validation scheme. Feature vectors for image texture are usually obtained from the analysis of groups of pixels and the way this analysis is performed is used to classify the different texture analysis methods. Five main categories can be distinguished: structural~\cite{Goyal1995,Serra1983,Zhang2002}, statistical~\cite{Haralick1979}, model-based~\cite{BackesCB09a,Backes2008,Cohen1991}, spectral~\cite{Chen1982,Tang1995}, and agent-based methods~\cite{BackesGMB10,Backes2010,Lai2006}. 

This paper proposes a novel method to analyze the structural elements of textures by means of a cellular automaton (CA) inspired by the pitting corrosion phenomenon, further on referred to as the Corrosion-Inspired Texture Analysis (CITA) method. The basic mechanisms behind this detrimental reaction which occurs between metals (or alloys) and their environment serve as inspiration to develop a CA-based model. Next, this CA-based model is employed to perform texture analysis by treating the image to be classified as a metal surface. The CITA method, like real corrosion, amplifies existing differences in material and height (in this case grayscale value) so that the biggest contrasts in the original texture image will become more pronounced and smaller contrasts will be nullified. The eroded mass of `metal' by the progression of pitting corrosion at each iteration is used to generate a feature vector that describes the image to be classified. The effectiveness of this strategy is demonstrated on two texture data sets with natural and synthetic textures. In addition, to verify the robustness of the CITA method, its invariance to noise and rotation were tested, obtaining satisfactory results. 

This paper is organized as follows. Section~\ref{sec:corrosion} describes the basics behind the pitting corrosion phenomenon, while the definition of a CA as well as further explanation of some parts of this definition form the subject of Section~\ref{sec:automata}. The CITA method is described in Section~\ref{sec:method} and the experimental setup needed to test its efficacy is explained in Section~\ref{sec:experiments}. Section~\ref{sec:resultsDiscussions} presents the results and discussion of the study. Finally, the paper is concluded in Section~\ref{sec:conclusions}.

%--------------------------------------------------------------------------------------
\section{Pitting corrosion} 
\label{sec:corrosion}

Corrosion is the disintegration of metals (and alloys) into their constituents due to reaction with the environment and is one of the main causes of structural failure in industrial systems, and poses as such an economic problem~\cite{roberge}. Dealing with corrosion is difficult because of its complex nature and the involvement of many variables. Therefore, modeling and simulation could allow for predicting more accurately the corrosion process in time. CA-based models are excellent candidates for modeling corrosion due to their intrinsic simplicity and therefore, since the beginning of the new millennium, attempts are being made to employ these models in the field of corrosion engineering~\cite{contreras,dicaprio,lishchuk,malki,pidaparti}. Corrosion is present in a wide range of metals and environments, which points to the universality of this phenomenon. The latter suggests that corrosion does not depend on the details of the underlying mechanism, so that it may be modeled adequately using simple models~\cite{valor}. Moreover, CA-based models are able to capture the stochasticity of the involved electrochemical reactions at the mesoscopic scale~\cite{lishchuk}.

\begin{figure}[!htbp]
\begin{center}
\includegraphics[width=\columnwidth]{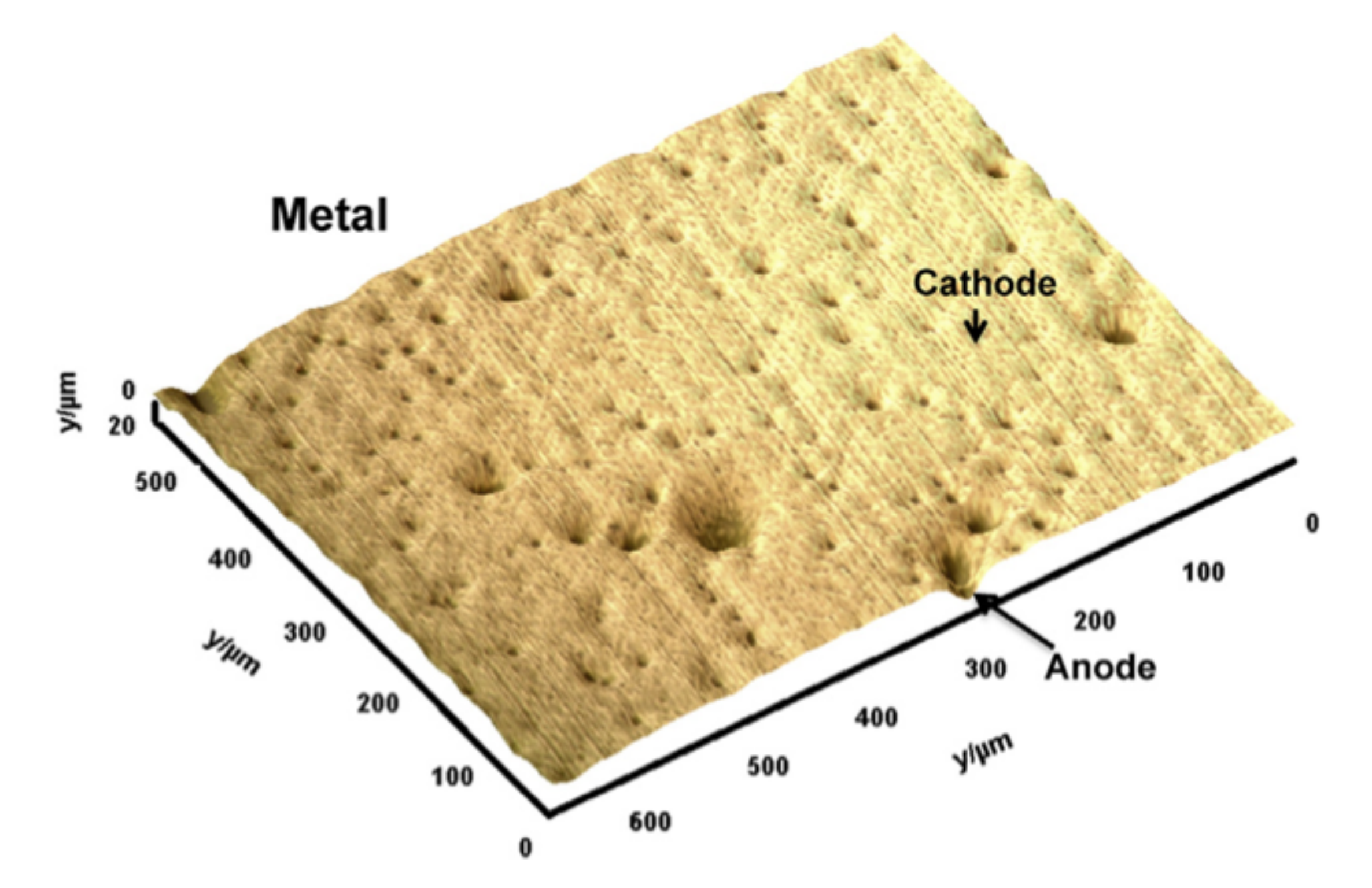}
\caption{Pitting corrosion: schematic representation in a metal surface.}
\label{fig:pitschema}
\end{center}  
\end{figure}

Pitting corrosion is a very harmful and common form of localized corrosion where all or most of the metal loss occurs concentrated in certain areas. Upon close inspection of the metal surface, pitting can be recognized by the appearance of small holes on the metal surface as shown in Figure~\ref{fig:pitschema}.%~\cite{roberge}. 
The first step in pitting corrosion is the pit initiation which is the result of impurities or irregularities of the metal surface or the environment, making perfectly polished surfaces more resistant to this type of corrosion. From there on, the acidity inside the pit is maintained by the spatial separation of the cathodic and anodic half-reactions, which creates a potential gradient and electromigration of aggressive anions into the pit (see Figure~\ref{fig:pitschema}). %~\cite{korb}. 
As pit growth progresses, different solution compositions develop inside the cavity and the consequent voltage (IR) drop along the metal/electrolyte interface illustrates that the deeper the pit, the lower the pit growth rate~\cite{malki,pickering,wang2}.

%\begin{figure}
%\centerline{
%\subfigure[]{
%\includegraphics[width=4cm]{pitting.pdf}
%\label{pitting}}
%\hfil
%\subfigure[]{
%\includegraphics[width=5.5cm]{pitschema.pdf}
%\label{pitschema}}}
%\caption{Pitting corrosion: (a) Metal surface and (b) schematic representation}
%\end{figure}

%--------------------------------------------------------------------------------------
\section{Cellular automata} 
\label{sec:automata}

CAs are mathematical constructs in which the space, state and time domains are discrete as opposed to partial differential equations (PDE) in which these three domains are continuous~\cite{berec,wolfram1}. The ability of CAs to generate a rich spectrum of sometimes complex spatio-temporal patterns from relatively simple underlying transition functions has led to their successful employment in the study of several (a)biological processes~\cite{milne,picioreanu,preziosi,schiff,vdweeen2,vasil}. Models based on CAs can be seen as an alternative to PDE-based models, to provide researchers with a wider range of modeling tools and, in some complex cases, a solution to problems encountered with some of the more classical modeling methods~\cite{yacoubi,toffoli}.

In this paper, we make use of a homogeneous CA, in which a single transition function, constructed using a combination of knowledge on the pitting corrosion phenomenon and intuition, governs the dynamics of all cells. The following definition of a homogeneous 2D CA is relied upon.
\npar
\textbf{Definition I}. \begin{em}(Homogeneous 2D cellular automaton)\label{def}\\
A homogeneous 2D cellular automaton $\mathcal{C}$ can be represented as
$$
\mathcal{C}=\left\langle \mathcal{T},S,s,N,\Phi\right\rangle\,,
$$
where 

\begin{enumerate}
\item[(i)] $\mathcal{T}$ is a two-dimensional grid of cells $c$.
\item[(ii)] $S$ is a finite set of $k$ states, with $S\subset\,\mathbb{N}$.
\item[(iii)] The output function $s$ yields the state $s(c,t)$ of every cell $c$ at the $t$-th discrete time step. 
\item[(iv)] The neighborhood function $N$ determines the neighboring cells of every cell $c$, including the cell $c$ itself.
\item[(v)] The transition function $\Phi$ yields the state $s(c,t+1)$ of every cell $c$ at the next time step, based on its state and that of its neighboring cells at the current time step.
\end{enumerate}
\end{em}

For reasons of comprehensiveness, some parts of this definition will be elaborated in the remainder of this section.

\subsection{Grid $\mathcal{T}$}

\begin{figure}[!htbp]
\begin{center}
\includegraphics[width=0.4\textwidth]{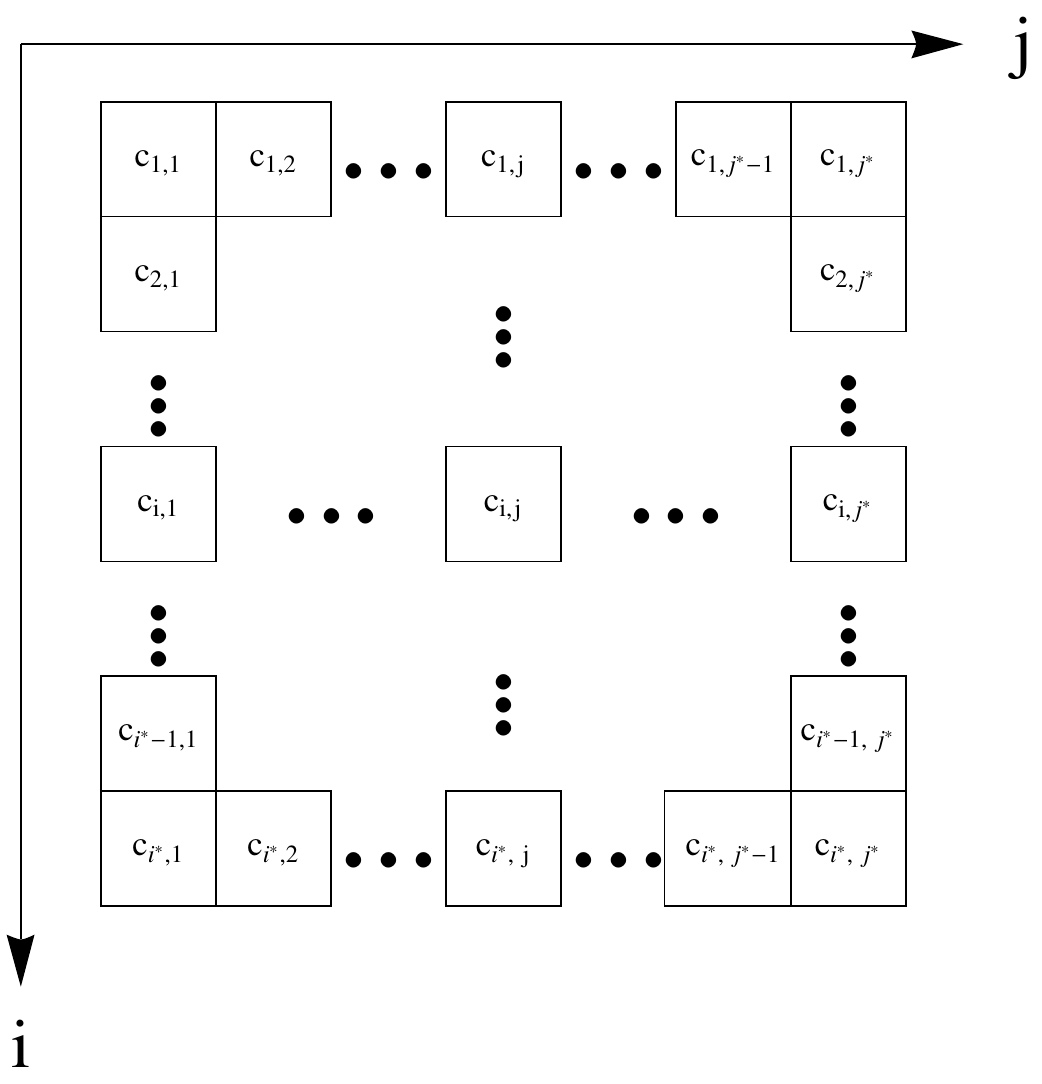}
\caption{Ordering of the cells of a 2D CA \label{cellen2D}}
\end{center}
\end{figure}

In this paper, a finite two-dimensional grid consisting of squares is used, because it has the most straightforward implementation and provides an easy way of linking the cells of $\mathcal{T}$ to the pixels of the texture images to be classified (cfr.\;infra). Furthermore, an indexing of the cells of a 2D CA is introduced, which is shown in Figure~\ref{cellen2D}. For a square grid, it holds that $i^*$ = $j^*$ = $\sqrt{\left|\mathcal{T}\right|}$.

\subsection{Neighborhood function $N$}

\begin{figure}[!htbp]
\centerline{
\subfigure[]{
\includegraphics[width=3cm]{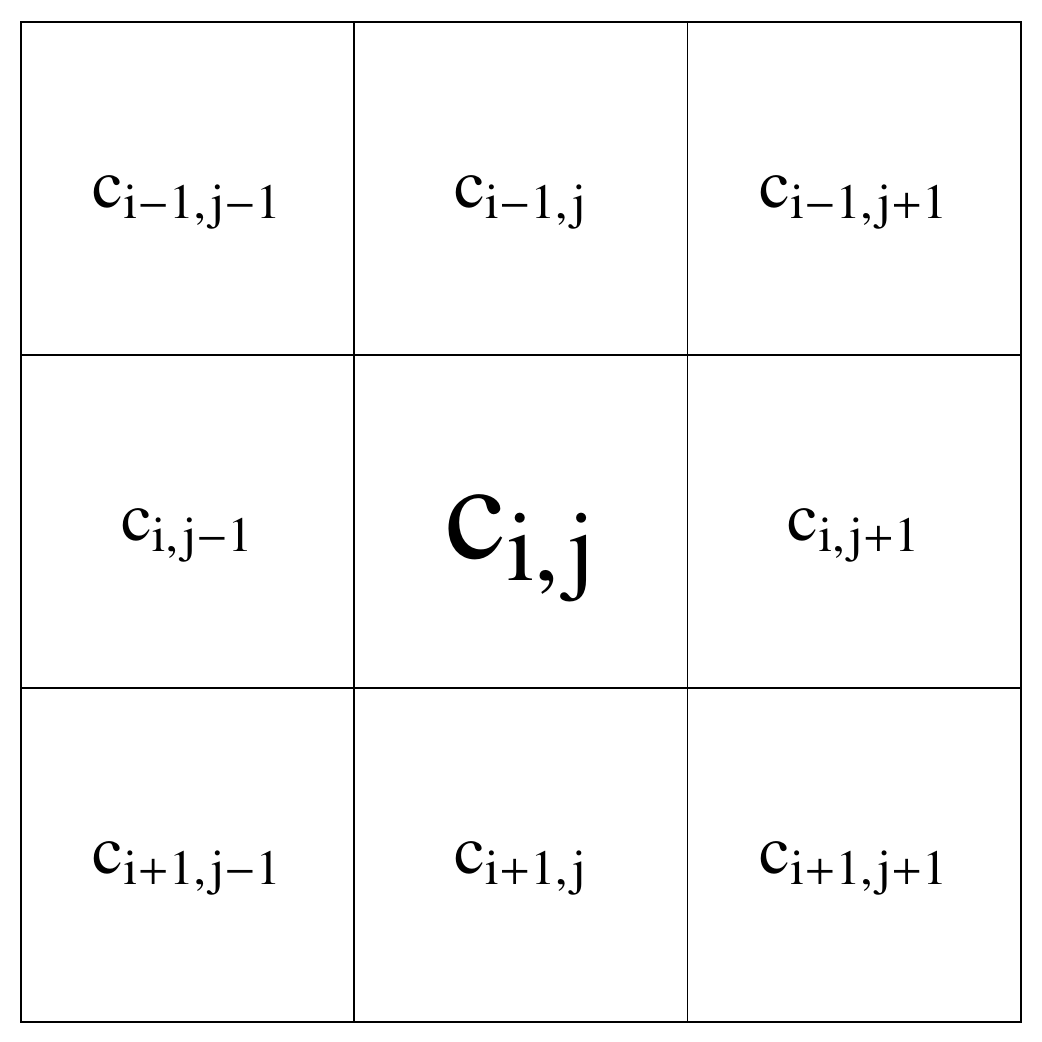}
\label{Moore}}
\hfil
\subfigure[]{
\includegraphics[width=3cm]{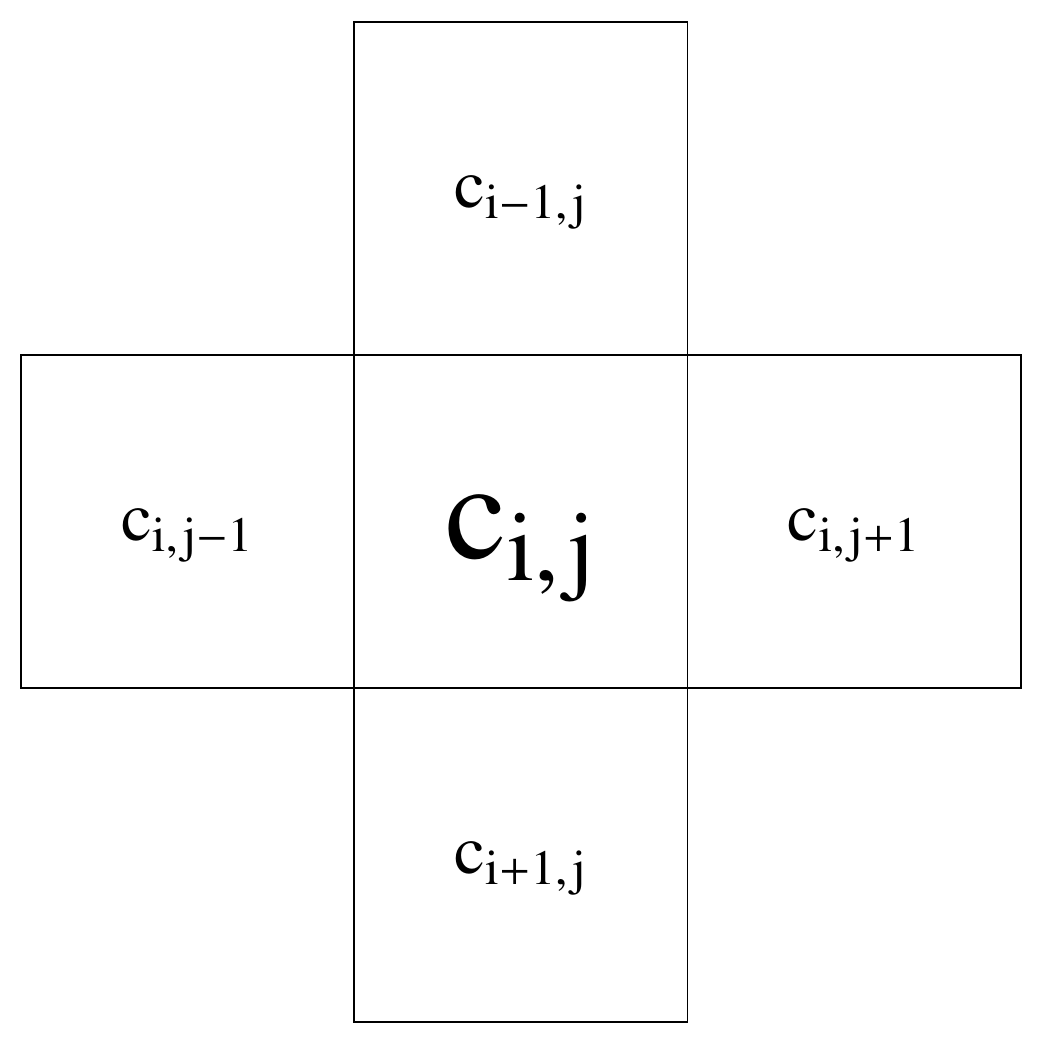}
\label{Neumann}}}
\caption{Neighborhoods of a cell $c_{i,j}$ in a square tessellation: (a) Moore neighborhood and (b) von Neumann neighborhood}
\end{figure}

Many different neighborhoods can be defined in 2D, the two most important ones being the Moore and the von Neumann neighborhood. The Moore neighborhood of a cell $c_{i,j}$ comprises those cells that share at least a vertex with $c_{i,j}$ (see Figure~\ref{Moore}). The von Neumann neighborhood is a more restricted neighborhood in which only those cells that share an edge with $c_{i,j}$ are considered as neighbors (see Figure~\ref{Neumann}). 

\subsection{Discrete states}

Every cell $c_{i,j}$ has one of the $k$ discrete states comprised in the set $S$. The states of the cells $c_{i,j}$ of $\mathcal{T}$ at $t=0$, i.e.\ $s(c_{i,j},0)$, constitute the initial condition of $\mathcal{T}$. In this paper, the initial condition of $\mathcal{T}$ is determined by the grayscale value of the different pixels of the corresponding texture image (cfr.\;infra).

\subsection{Transition function $\Phi$}

The transition function $\Phi$ determines the state of a cell $c_{i,j}$ at the ($t + 1$)-th time step based on the cell's current state and the states of its neighboring cells. The transition function employed in this paper is executed in a deterministic and synchronous manner, meaning that $\Phi$ is used to evaluate the state of every cell of $\mathcal{T}$ at every time step and for all cells at the same time~\cite{baetvdw}.

%--------------------------------------------------------------------------------------
\section{Corrosion-Inspired Texture Analysis}
\label{sec:method}

The CITA method proposed in this paper starts by converting the texture image into the initial state of a CA. Thereafter, a CA-based model inspired by the pitting corrosion phenomenon is evaluated for a number of time steps. The cumulative mass of corroded metal after each iteration of the CA-based model is used to construct a feature vector for every texture image. Finally, these vectors of characteristics are used to classify the images via LDA. In the remainder of this section, the CITA method is explained in more detail.

A two-dimensional grayscale image is treated as a discrete object and is seen as a grid $\mathcal{T}$. The dimensions of $\mathcal{T}$ are defined by the size of the image, where each pixel of the image is a cell of the CA. The original image is then used to determine an initial state of the cells of the CA by converting the gray level image into a discrete initial state for each cell. Thus, for the initial configuration $s(c_{i,j},0)$ there are 256 possible states, ranging from 0 to 255. This conversion is described by 

\begin{equation}
	s(c_{i,j},0) = I(i,j)\,,
	\label{conversion}
\end{equation} 
where $I$ is the original image and $I(i,j)$ represents the gray level of the pixel at the $i$-th row and $j$-th column of the image $I$. In order to introduce the ideas of pitting corrosion, the 2D grid will be regarded as a metal surface and the state of each cell will represent the depth of the local pit in the metal (i.e.\ along the third dimension), with state 0 meaning that there is no pit and 255 being the largest pit depth of the metal at $t = 0$. It is important to point out that for $t > 0$ the maximum pit depth can exceed 255 and from thereon it is possible that the grid can no longer be represented as a grayscale image.

An important consideration is the choice of boundary conditions in order to obtain an appropriate behavior of the CA-based model. The two most popular boundary conditions are the periodic and reflecting boundary conditions. The former tries to simulate an infinite grid, where the new boundaries of the top, bottom, left and right are filled with the values of the opposite side, thus forming a torus in a 3D space. This boundary condition is useful for simulating systems where the physical boundaries do not play an important role. However, throughout this paper, reflecting boundaries will be used as they give rise to better results for the studied databases as was observed from preliminary tests. Firstly, an imaginary row at the top and at the bottom of the grid and an imaginary column at the left and at the right of the grid are added. Then, the reflecting boundary conditions are applied at every time step as follows:
\begin{eqnarray}
s(c_{1,j},t) &=& s(c_{2,j},t),\nonumber \\
s(c_{n+2,j},t) &=& s(c_{n+1,j},t),\nonumber \\
s(c_{i,1},t) &=& s(c_{i,2},t),\nonumber \\
s(c_{i,n+2},t) &=& s(c_{i,n+1},t),
\label{eq:reflecbound}
\end{eqnarray}
with $n$ the size of the original image with $n \times n$ pixels.

The updated state of each cell $c_{i,j}$ of $\mathcal{T}$ at time $t+1$ depends on the analysis of the states of the cells in the neighborhood of $c_{i,j}$ at time $t$. In this paper, the Moore neighborhood (see Figure~\ref{Moore}) is employed. Furthermore, the CITA method makes use of a transition function $\Phi$ inspired by pitting corrosion. In a first step, $d_{i,j}$ is calculated for every cell $c_{i,j}$ as the difference between the state of this cell and the lowest state value within its Moore neighborhood (see Eq.\;(\ref{eq:dif})): 

\begin{equation} 
	\label{eq:dif}
	d_{i,j} = s(c_{i,j},t) - \text{min} (\tilde{s}(\textit{N}(c_{i,j}),t))\,,
\end{equation}
where $\tilde{s}(\textit{N}(c_{i,j}),t)$ is the set of states of the cells in the Moore neighborhood of $c_{i,j}$.

Bearing in mind the principles of pitting corrosion, a local `impurity' or minimum height difference is needed at a certain location in order to initiate or propagate pitting corrosion. For this purpose, a surface roughness parameter $\nu$ is introduced. All differences lower than this parameter $\nu$ are considered insignificant, i.e. not real impurities, in order to account for the fact that not even a polished metal surface is perfectly smooth. This means that differences $d_{i,j}$ lower than $\nu$ will not give rise to (further) pitting. On the other hand, the larger the difference grows, the lower the pit growth rate will be due to the IR drop, until finally the pit growth rate becomes zero. In this paper, it is assumed that if the difference $d_{i,j}$ is greater than 254, the greatest possible difference at $t=0$, the corresponding pit growth rate is zero. This means that only the state of those cells with a difference $d_{i,j}$ greater than or equal to $\nu$ and smaller than 255 are evaluated.

\begin{figure}[!htbp]
\centerline{
\subfigure[]{
\includegraphics[width=2.5cm]{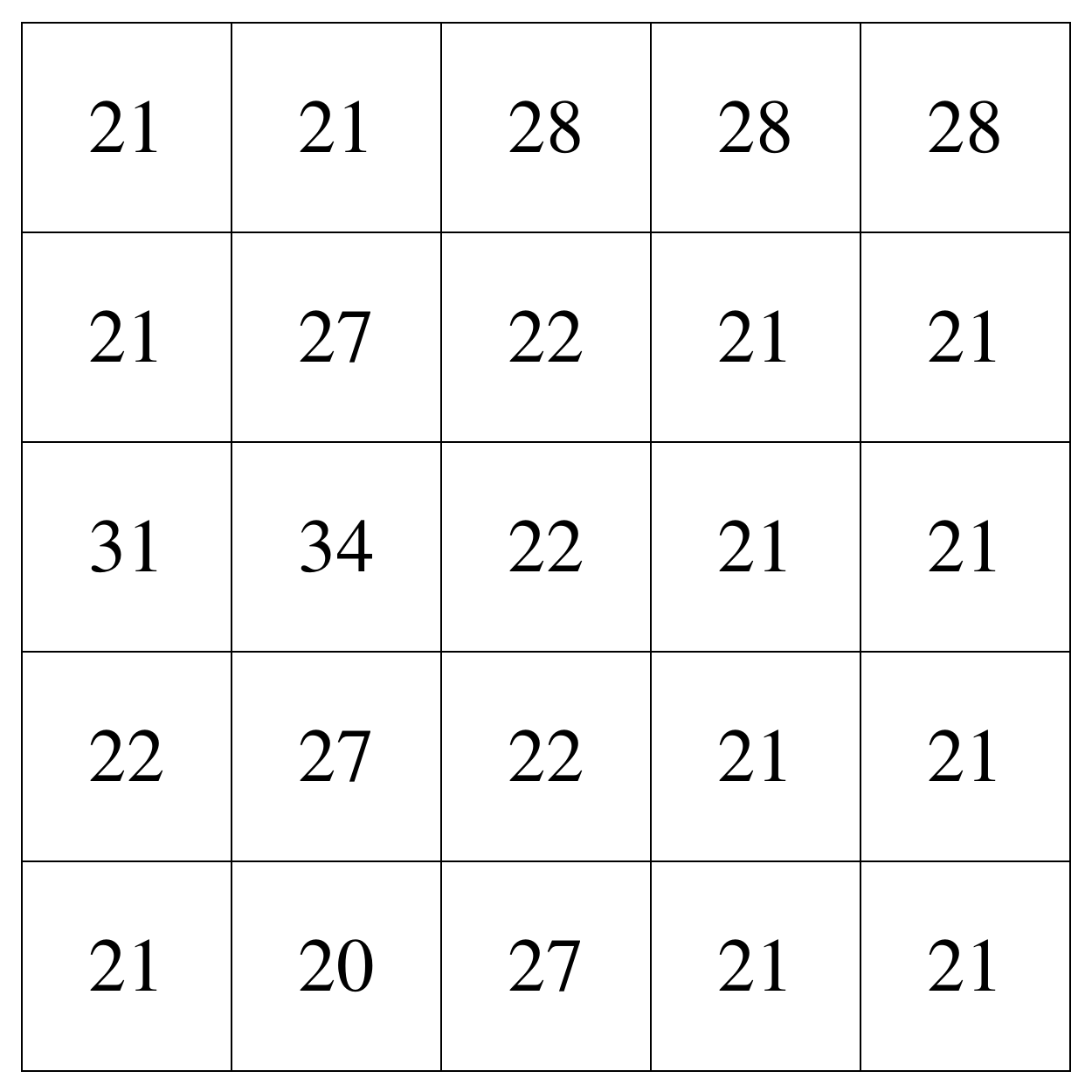}
\label{fig:dif1}}
\hfil
\subfigure[]{
\includegraphics[width=2.5cm]{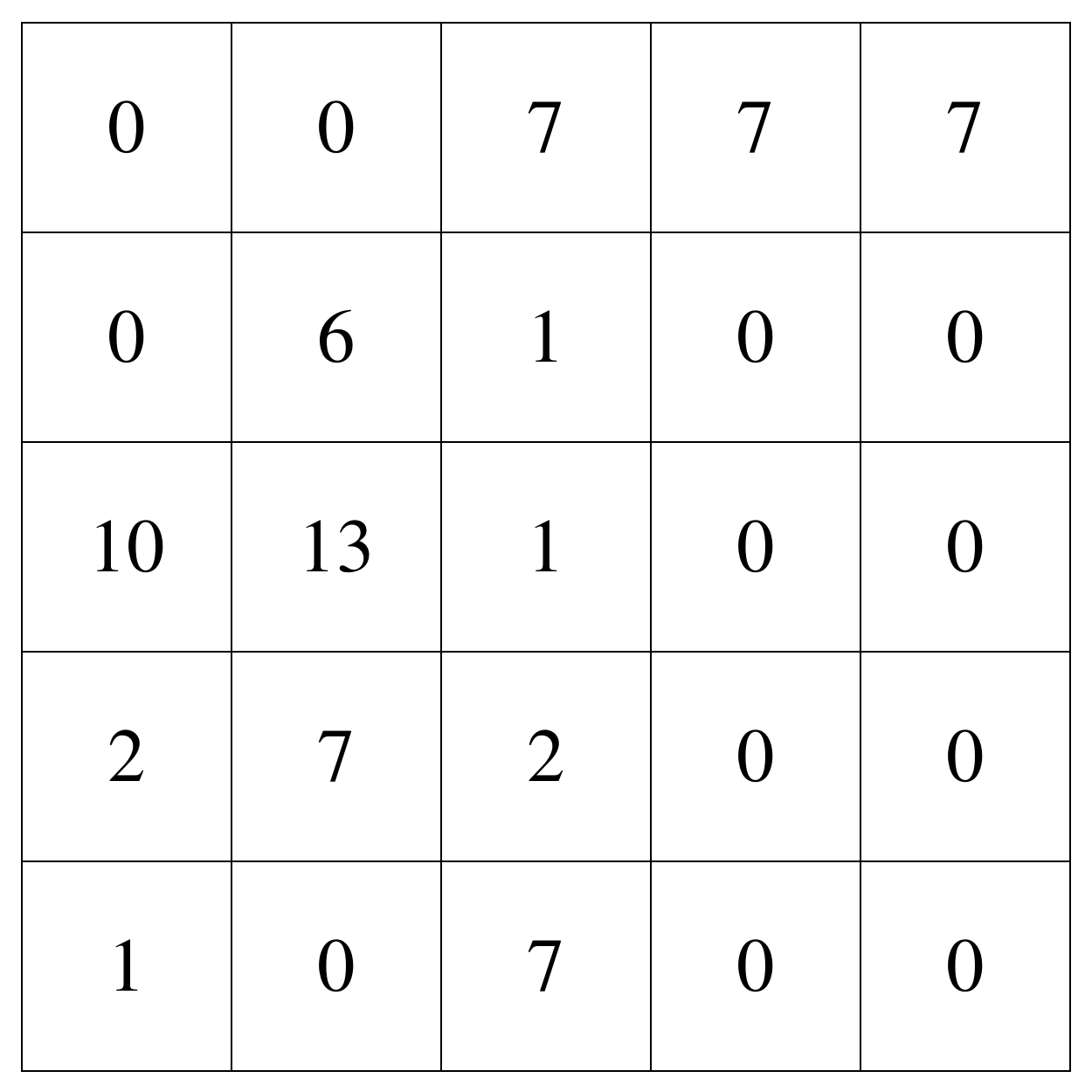}
\label{fig:dif2}}
\hfil
\subfigure[]{
\includegraphics[width=2.5cm]{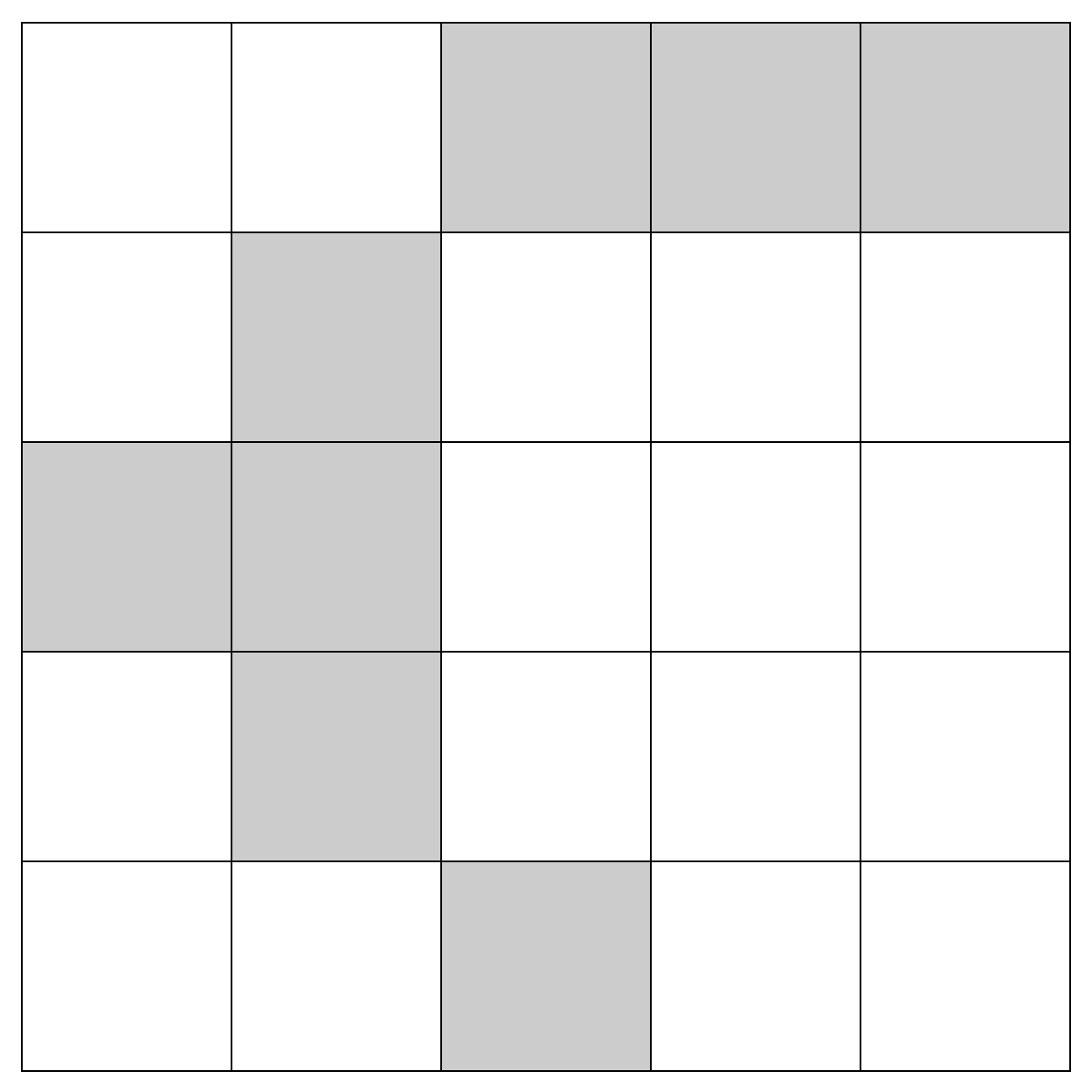}
\label{fig:dif3}}}
\caption{Selection of cells to be updated, with $\nu = 5$: (a) $5 \times 5$ square grid with initial states of the cells, (b) difference $d_{i,j}$ for all cells according to Eq.\;(\ref{eq:dif}) and (c) gray cells indicate cells to be updated}
\end{figure}

Figures~\ref{fig:dif1}-\ref{fig:dif3} illustrate the selection process to determine whether a cell will be evaluated or not. In this example, $\nu$ is set to five meaning that cells $c_{i,j}$ whose state differs less than five with the lowest state in its neighborhood are considered to belong to the surface and will not have their state changed. Figure~\ref{fig:dif1} shows the cells belonging to a $5 \times 5$ square tessellation with their initial state. Figure~\ref{fig:dif2} depicts the difference $d_{i,j}$ for each of these cells calculated according to Eq.\;(\ref{eq:dif}). Finally, in Figure~\ref{fig:dif3} the gray cells indicate the cells that are evaluated in that time step, because their $d_{i,j}$ is greater than or equal to five and smaller than 255.

Under these assumptions, the transition function $\Phi$ establishes the state of a cell $c_{i,j}$ at the ($t + 1$)-th time step according to

\begin{equation}
s(c_{i,j},t+1) = 
\begin{cases}
s(c_{i,j},t) + Q(d_{i,j}, \gamma) & ,\,\text{if } 255 > d_{i,j} \geq \nu,\\
s(c_{i,j},t) & ,\,\text{if } d_{i,j} < \nu \text{ or } d_{i,j} \geq 255,
\label{eq:cases}
\end{cases}
\end{equation}
where $\gamma \in [0,1]$ is the pitting power. This parameter $\gamma$ represents the metal-specific resistance to corrosion under given environmental conditions, where $\gamma = 0$ stands for completely resistant metal. Further, $Q$ is a function that employs $d_{i,j}$ and $\gamma$ to determine the level of corrosion to be applied. In this paper, $Q$ is defined as

\begin{equation}
	\label{eq:T}
	Q(d_{i,j}, \gamma) = (255-d_{i,j})\,\gamma.
\end{equation}

From Eq.~(\ref{eq:T}), it can be seen that the function $Q$ gives, depending on the value of $\gamma$, a non-integer output, meaning that the employed model structure is actually a continuous CA or Coupled Map Lattice rather than a CA~\cite{levine}. However, in order to keep working with a CA-based model and to not overcomplicate the model, the choice was made to limit the output of $Q$ to integer values (see Eq.~(\ref{eq:T2})).

\begin{equation}
	\label{eq:T2}
	Q(d_{i,j}, \gamma) = \left\lfloor (255-d_{i,j})\,\gamma\right\rfloor,
\end{equation}
where $a$ in $\left\lfloor a \right\rfloor$ denotes the floor of $a$.

The output of the CA-based model at every time step is the cumulative mass of corroded product. In each iteration, after updating the state of the cells, the mass that suffered corrosion in that iteration is added to the eroded total mass from the previous iteration. Finally, this cumulative corroded mass is expressed relative to the number of pixels of the texture images such that texture images with different sizes can be compared using the CITA method.

%In essence, for cell $c_{i,j}$ of $\mathcal{T}^*$ at the $t$-th time step, the state of the neighbors is determined. Next, the neighbor with the lowest state is used to calculate the difference $d_{i,j}$. If this difference is less than the chosen value of the surface roughness $\nu$ or more than 254, the state of the cell remains the same at $t + 1$, but if these restrictions are not met, the state of the cell will be updated according to $T$ (see Equation~\ref{eq:T}).

\begin{figure}[!htbp]
\begin{center}
  \includegraphics[width=\columnwidth]{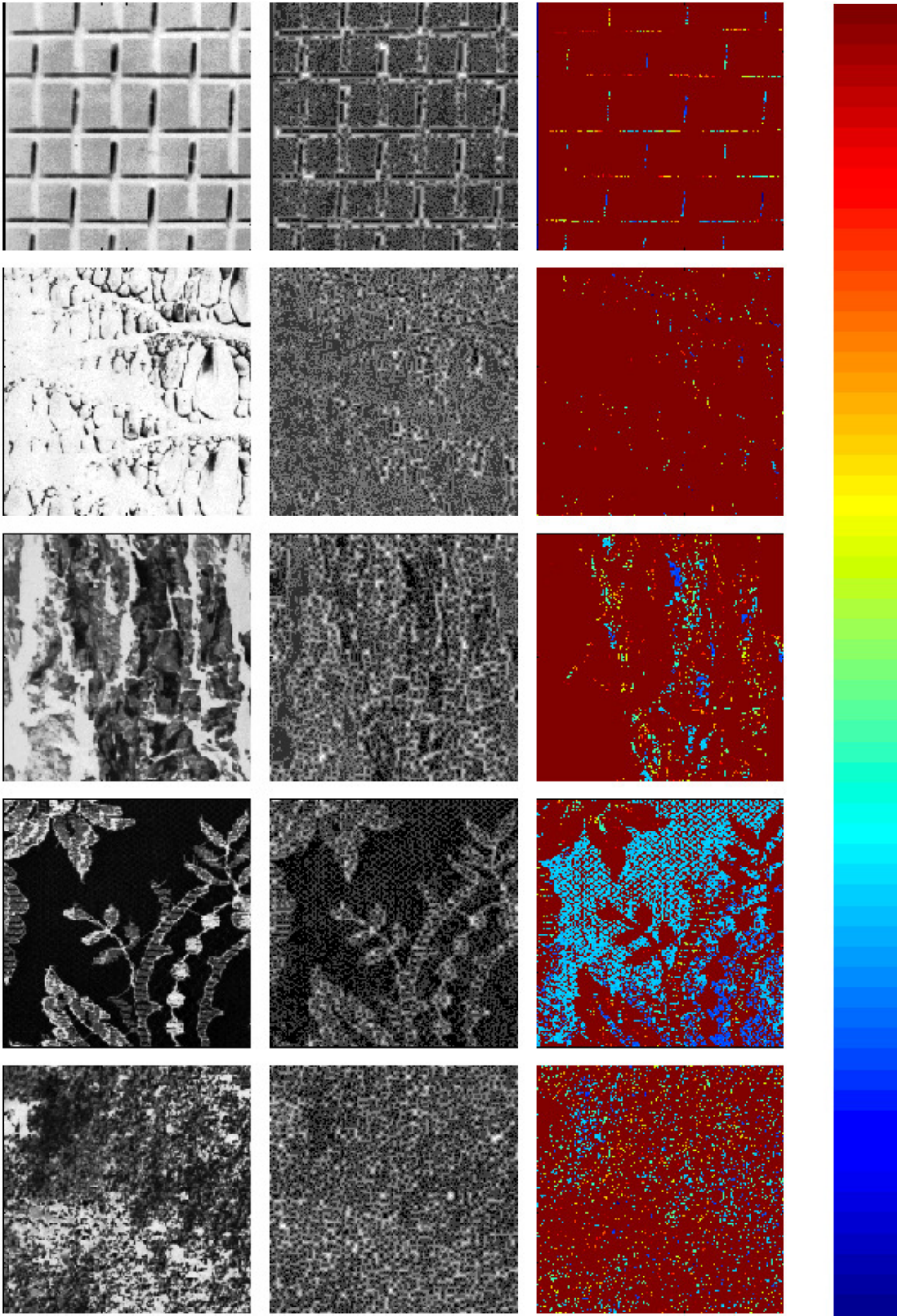}
\caption{Simulation results. First column: original images. Second column: result in grayscale after application of CA-based model. Third column: result in blue-red scale after application of CA-based model (see color code at the right hand side). The experiments were performed with $\gamma$ = 0.05, $\nu$ = 5 and 90 iterations.}
\label{fig:cabased}
\end{center}  
\end{figure}

Figure~\ref{fig:cabased} shows some examples of initial images and the simulated result after 90 iterations of the pitting-corrosion-inspired CA-based model. In the first column the original images are shown, the second column shows the simulated output of the CA-based model in grayscale, while the third column displays the same results as the second column, but scaled in a color map where blue indicates the lowest and red the highest resulting values. After simulation, some structural details from the original image can still be retrieved in the simulated output. Regions with similar state values are mostly considered by the model as belonging to the same local surface and therefore tend to keep the same state value throughout the simulation.

\begin{figure}[!ht]
\begin{center}
  \includegraphics[width=\columnwidth]{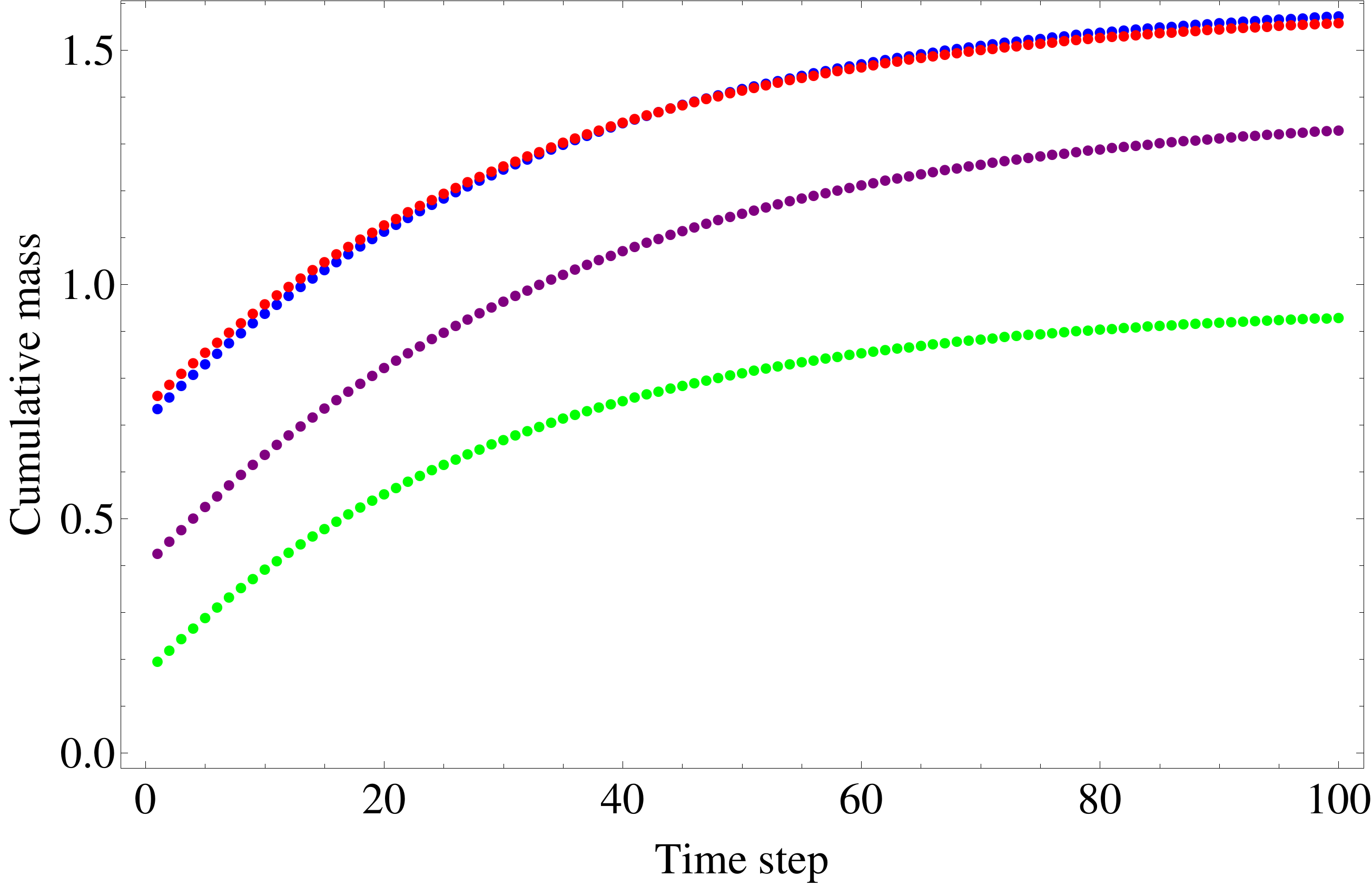}
\caption{Feature vectors from four different texture images of the Brodatz database, with $\gamma$ = 0.04, $\nu$ = 5 and 100 iterations. The blue and red vectors originate from class 11 images, the purple vector from a class 24 image and the green vector from a class 78 image.}
\label{fig:featurevector}
\end{center}  
\end{figure}

In a final step, the time series of cumulative mass of corroded metal, relative to the total number of pixels, is used as the feature vector for each texture image and classification is performed using LDA following a stratified 10-fold cross-validation scheme. LDA is traditionally used in texture analysis to find a linear combination of attributes resulting in a good separation of the classes. The proposed method is summarized in Figure~\ref{algCITA}. Figure~\ref{fig:featurevector} shows the feature vectors for four texture images from the Brodatz database. It can clearly be seen from this figure that the feature vectors from textures belonging to the same class are very similar on the one hand and that these vectors are different from the feature vectors from images belonging to different classes on the other hand.

\begin{figure}[!htbp]
\begin{algorithmic}
%\begin{boxedalgorithmic}
\State \textbf{Input:} Original image ($n \times n$), surface roughness $\nu$, pitting power $\gamma$, number of iterations
\State \textbf{Output:} Class label
\npar
\State $s(c_{i,j},0) \leftarrow$ original image ($I(i,j)$)
\npar
\State \emph{Add imaginary row at top and bottom}
\State \emph{Add imaginary column at left and right}
\npar
\ForAll{iterations}
\npar
\State \emph{Apply boundary reflection on grid $\mathcal{T}$:}
\npar
\State \hspace{0.3cm} $s(c_{1,j},t) = s(c_{2,j},t)$
\State \hspace{0.3cm} $s(c_{n+2,j},t) = s(c_{n+1,j},t)$
\State \hspace{0.3cm} $s(c_{i,1},t) = s(c_{i,2},t)$
\State \hspace{0.3cm} $s(c_{i,n+2},t) = s(c_{i,n+1},t)$
\npar
\For{$i = 2 \to n+1$} 
\For{$j = 2 \to n+1$} 
\State $d_{i,j} \leftarrow s(c_{i,j},t) - \text{min}(\tilde{s}(\textit{N}(c_{i,j}),t))$	   			 	
\If{($d_{i,j} < \nu\; \text{or}\; d_{i,j} \geq 255$)}
\State $s(c_{i,j},t+1)  \leftarrow s(c_{i,j},t)$
\Else
\State $s(c_{i,j},t+1)  \leftarrow s(c_{i,j},t) + Q(d_{i,j},\gamma)$
\EndIf
\EndFor
\EndFor
\npar
\State \emph{Calculate cumulative corroded mass}
\npar
\EndFor
\npar
\State \emph{Feature vector} $\leftarrow$ \emph{cumulative corroded masses relative to number of pixels}
\State \emph{Do classification via LDA}
%\end{boxedalgorithmic}
\end{algorithmic} 
\caption{Pseudocode for the CITA method.}
\label{algCITA}
\end{figure}

%--------------------------------------------------------------------------------------
\section{Experimental setup}
\label{sec:experiments}
 
To validate the CITA method, it is employed for the classification of the images of two classical texture databases, the Brodatz and Vistex databases, and the results obtained by applying the CITA method are compared with the results obtained with several established methods from literature. The remainder of this section includes a short description of the employed databases and the methods from literature used for comparison as well as an optimization of the parameters of the CITA method, i.e.\ $\gamma$, $\nu$ and the number of iterations performed in order to obtain the most favorable results. To ensure the CITA method is not sensitive to the parameter configuration, Usptex, a different database than the databases used for the validation of the method is used to perform the parameter optimization. 

%--------------------------------------------------------------------------------------
\subsection{Databases}

Two important databases, widely used in the literature and each with its own peculiarities, are employed for testing the different methods for pattern classification: the Brodatz and Vistex databases and a third database, Usptex, is used to perform the parameter optimization.

\subsubsection{The Brodatz database~\cite{Brodatz66}} 
\begin{figure}[!htbp]
\begin{center}
  \includegraphics[width=\columnwidth]{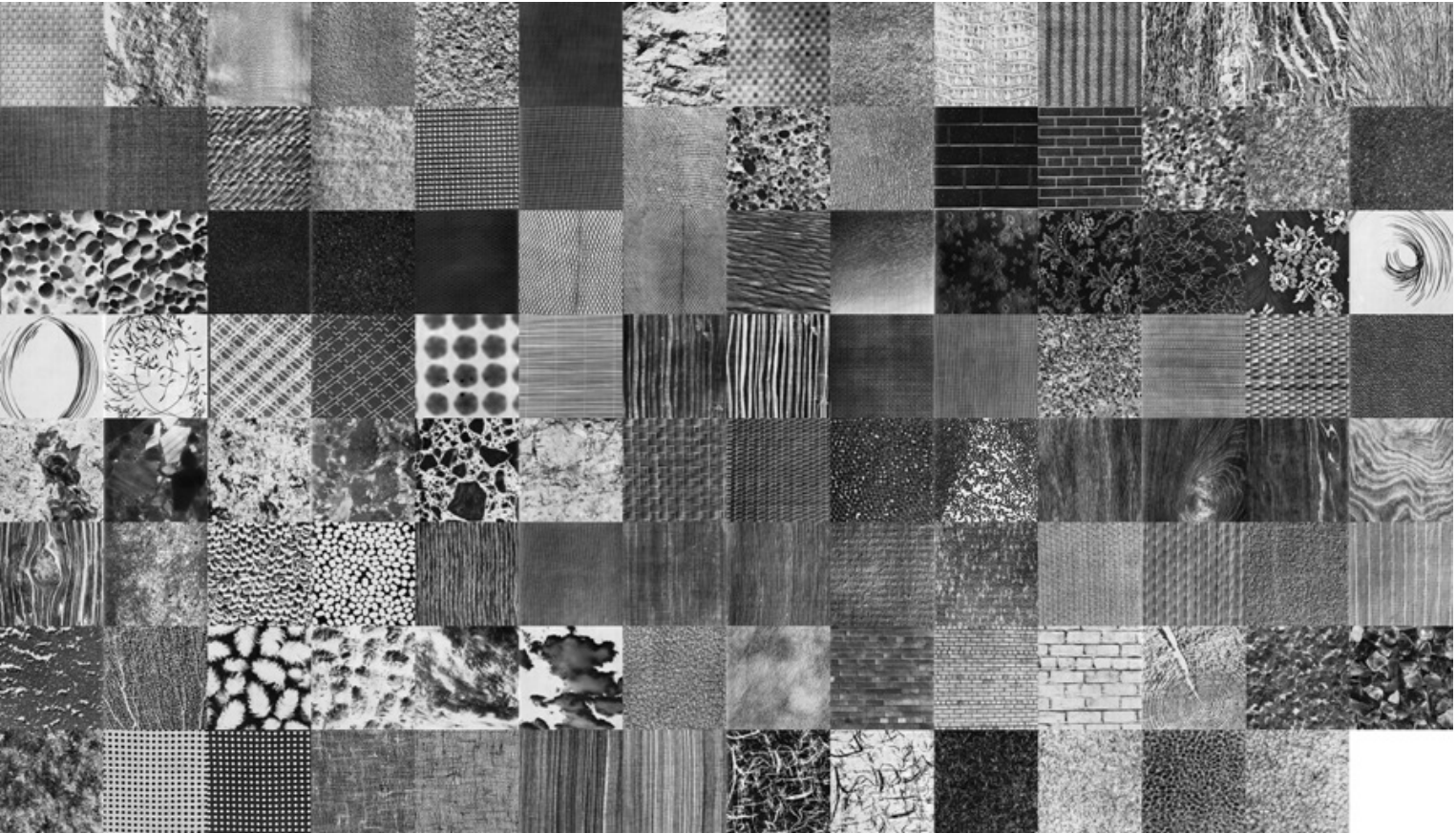}
	\caption{The Brodatz database. The size of the images is $640 \times 640$ pixels.}
\label{fig:brodatz}  
\end{center}     
\end{figure}

\begin{figure*}[!htbp]
\centerline{
\subfigure[]{
\includegraphics[width=3cm]{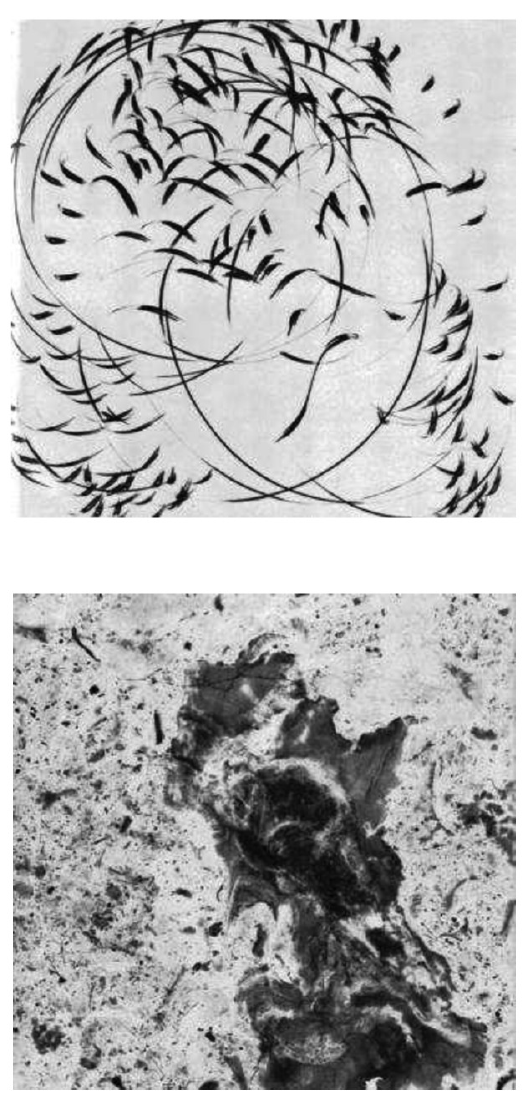}
\label{fig:subimagesa}}
\hfil
\subfigure[]{
\includegraphics[width=7.5cm]{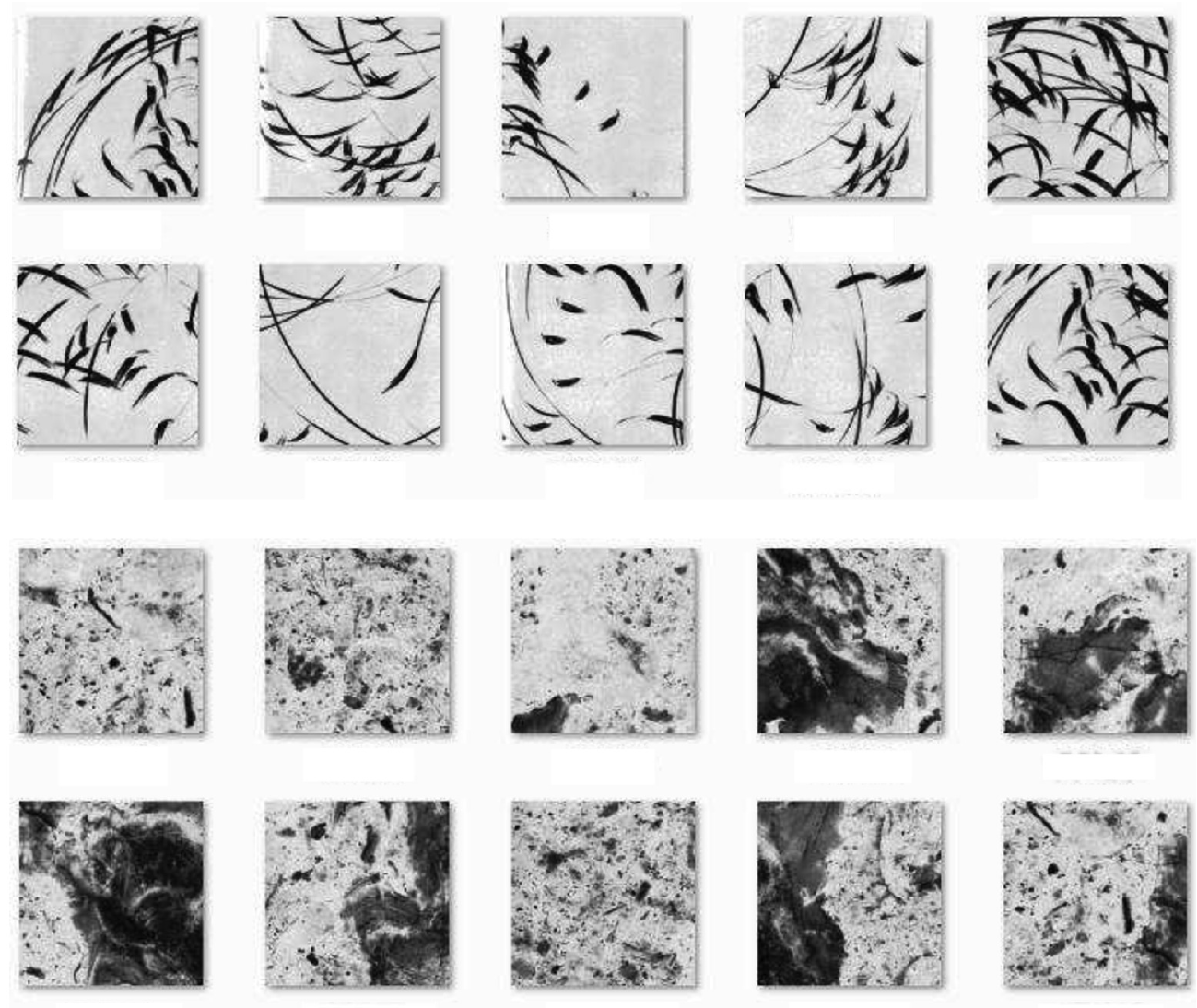}
\label{fig:subimagesb}}}
\caption{(a) Two Brodatz textures with a size of $640 \times 640$ pixels and (b) ten subimages with a size of $200 \times 200$ pixels.}
\end{figure*}

contains 111 unique natural textures (and therefore also 111 classes) with image size of $640 \times 640$ pixels and 256 gray levels. From each image ten subimages with size of $200 \times 200$ pixels were obtained, resulting in an image database containing 1110 images. Figure~\ref{fig:brodatz} shows the complete Brodatz database and Figures~\ref{fig:subimagesa}~and~\ref{fig:subimagesb} show for two of the original Brodatz images the ten selected subimages.  

\subsubsection{The Vistex database~\cite{Vistex09}} 
\begin{figure}[!htbp]
\begin{center}
  \includegraphics[width=0.7\columnwidth]{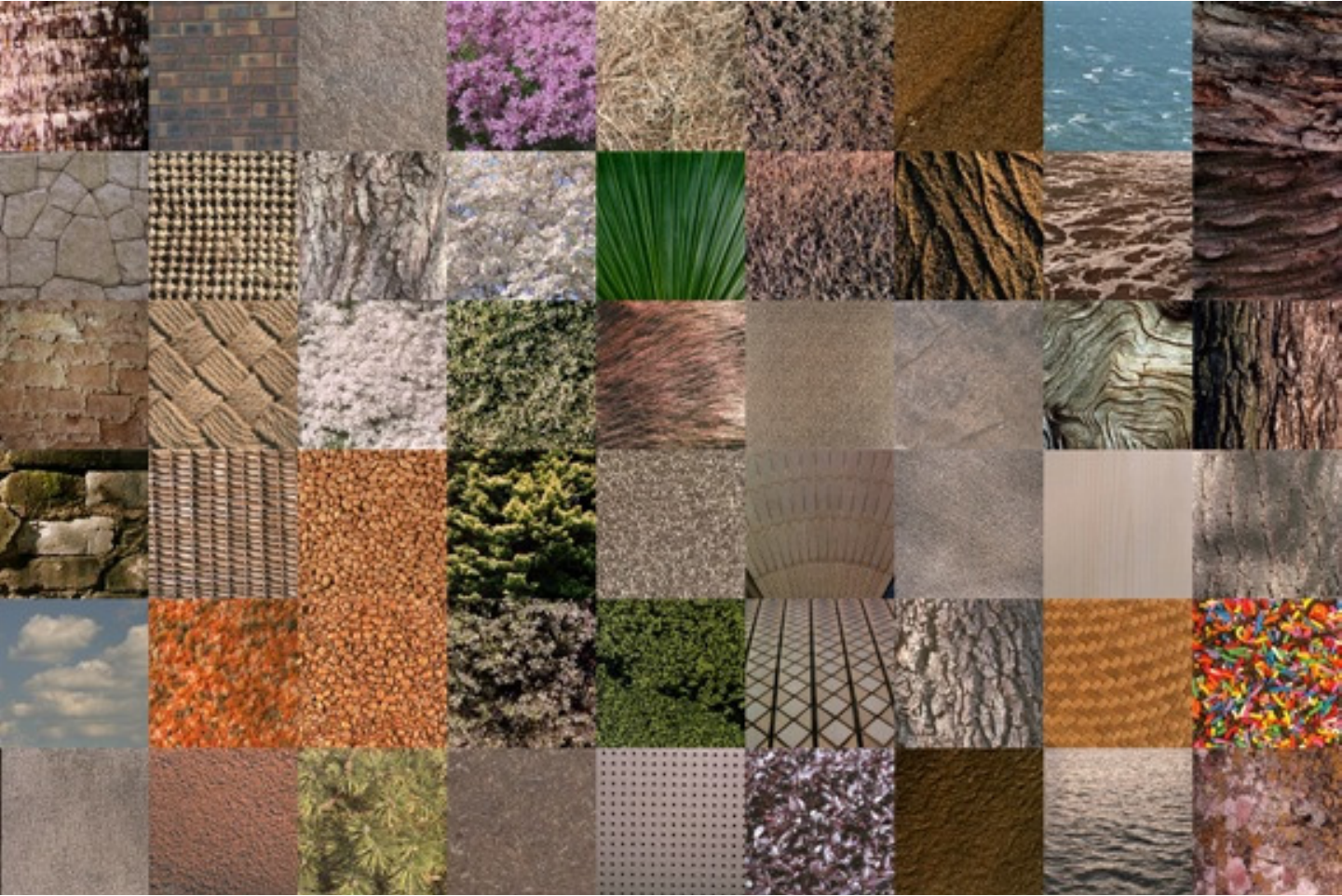}
	\caption{The Vistex database. The size of the images is $128 \times 128$ pixels.}
\label{fig:vistex}  
\end{center}     
\end{figure}

contains 864 images belonging to 54 texture classes. Each texture class contains 16 texture samples of $128 \times 128$ pixels, each extracted from a particular texture pattern without overlapping (see Figure~\ref{fig:vistex}). The true color RGB images are converted to grayscale intensity images, because the CITA method in its present form works only on grayscale images.
%ftp://whitechapel.media.mit.edu/pub/VisTex
 
\subsubsection{The Usptex database~\cite{Backes2012}} 
\begin{figure}[!htbp]
\begin{center}
  \includegraphics[width=\columnwidth]{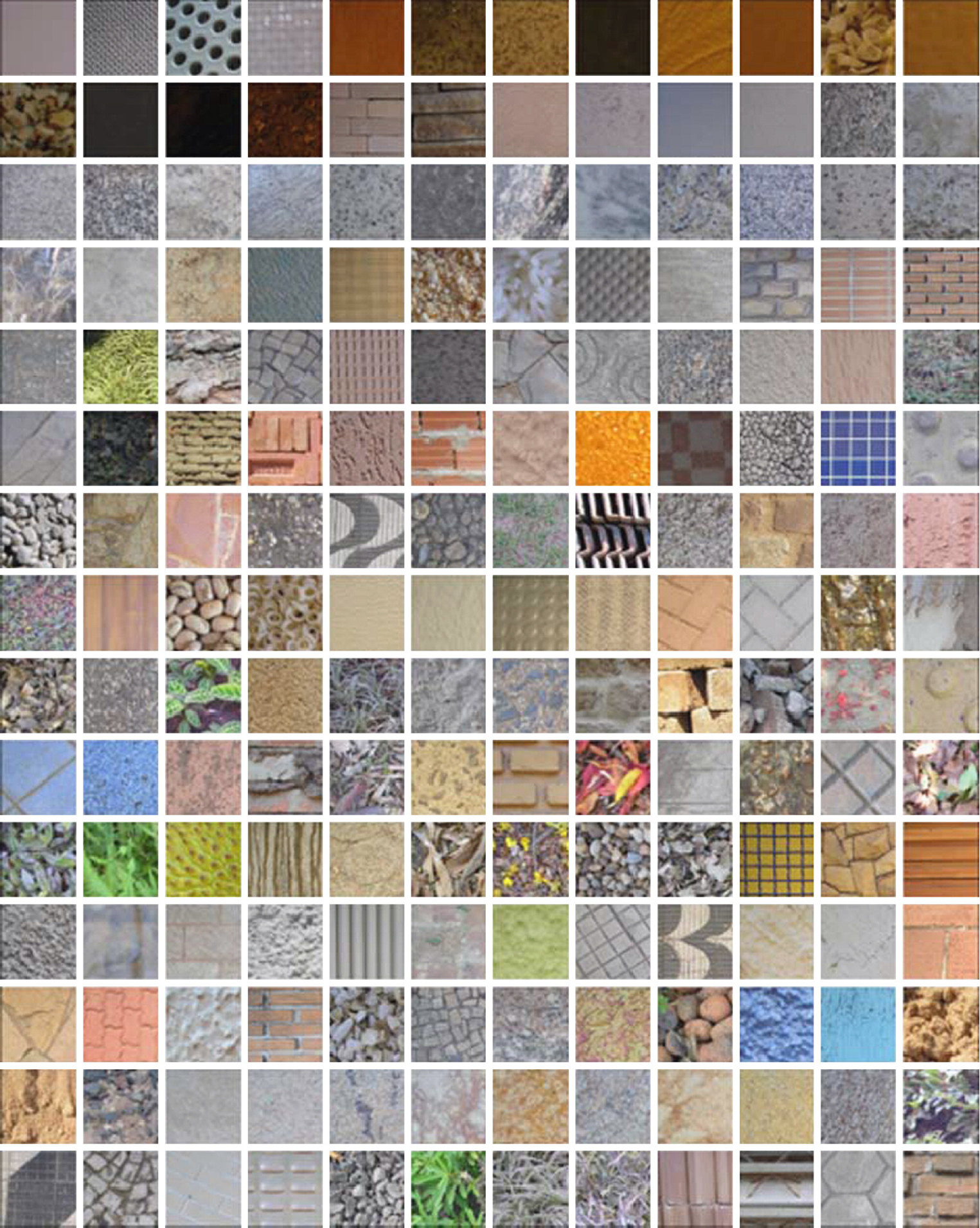}
	\caption{The Usptex database. The size of the images is $512 \times 384$ pixels.}
\label{fig:usptex}  
\end{center}     
\end{figure}

contains 191 color images that each form a texture class (see Figure~\ref{fig:usptex}). Each image has a size of $512 \times 384$ pixels from which 12 subimages with a size of 128 $\times$ 128 pixels are extracted without overlapping, so that a total of 2292 images is obtained. The images are again converted to grayscale images.

%--------------------------------------------------------------------------------------
\subsection{Established methods for texture analysis}

\subsubsection{Fourier descriptors~\cite{Azencott1997,Gonzalez2006}} consider attributes in terms of spectral density considering the texture as a Gaussian random field. The Fourier transform was calculated for each image, where the spectrum was divided into 64 sectors with eight radial distances and eight angles. The sum of the absolute spectrum values for each sector is calculated, resulting in 64 descriptors per image.

\subsubsection{Gray Level Co-occurrence Matrix - GLCM~\cite{Haralick1979}} is based on the spatial gray level dependence matrices. Haralick descriptors (Contrast, Correlation, Energy and Homogeneity) were computed from resulting co-occurrence matrices with angles of 0$^{\circ}$, 45$^{\circ}$, 90$^{\circ}$ and 135$^{\circ}$, distances equal to one or two pixels and 64 gray levels in order to obtain a set of 32 descriptors for each image.

\subsubsection{Gray Level Difference Matrix - GLDM~\cite{Weszka1976,Kim1999}} calculates the absolute gray level difference between two pixels with distance $h$. Here, 60 descriptors were obtained using $h =$~1, 3 and 5 and the attributes contrast, angular second moment, entropy, mean, and inverse difference moment from the estimated probability density function.

\subsubsection{Gabor filter~\cite{Manjunath1996,Daugman1998,Idrissa2002}} is a bi-dimensional Gaussian function modulated with an oriented sinusoid in a determined frequency and direction. To perform the tests, 64 filters were used, composed of eight rotation filters and eight scale filters with lowest frequency equal to 0.01 and highest frequency equal to 0.4.

\subsubsection{Local Binary Pattern Variance - LBPV~\cite{Guo2010}} is a variation of traditional LBP~\cite{Ojala2002} and is calculated from the binary value of each pixel in the radius 1 neighborhood surrounding the central pixel, measuring the local variance. 

%--------------------------------------------------------------------------------------
\subsection{Parameter Evaluation} 

The parameterization of the CITA method is performed using the Usptex database, a different database than the ones used for validation. This is done to ensure that the CITA method is not susceptible to the parameters and therefore the same configuration can be used for classification of textures from different databases. 

\subsubsection{Number of iterations} 

To describe each image, the cumulative mass of corroded metal after each iteration of the CA-based corrosion model is used. These values constitute the vector of characteristics that is used to discriminate each of the images. However, finding a single number of iterations that gives rise to the smallest, most informative feature vector for all images of both databases is a non-trivial task due to the variety of the type of texture images and also because this number is dependent on the values of $\gamma$ and $\nu$. In order not to overcomplicate the problem, the choice is made to look for a single optimal number of iterations for both databases that overall gives the best result for all the texture images in both databases. This optimal number of iterations is nevertheless still kept dependent on $\gamma$ and $\nu$. 

\subsubsection{Surface roughness $\nu$}

One of the parameters that defines the pitting corrosion is the surface roughness $\nu$. According to the proposed corrosion-based method, pixels having a difference $d$ lower than $\nu$ (Eqs.~(\ref{eq:dif})~and~(\ref{eq:cases})) do not suffer from the action of the corrosion process, considering that they are part of the local surface. However, if the neighborhood has a difference $d$ greater than the permitted threshold surface, the center pixel will pass through a corrosion process having its value eroded according to Eqs.~(\ref{eq:cases})~and~(\ref{eq:T}).  Figure~\ref{fig:parametersusptex} shows the success rate surface, i.e.\ the percentage of correctly classified texture images, for the Usptex database for $\nu$ varying from 0 to 10 and and $\gamma$ varying from 0.01 to 0.08. The figure shows that higher values of $\nu$ lead to a lower success rate. However, when $\nu$ equals 0 the obtained success rate is smaller than for $\nu$ equal to 1 for almost all values of $\gamma$. Thus, a value of 1 for $\nu$ is chosen as optimal value.

\subsubsection{Pitting power $\gamma$}

Another model parameter with a physical meaning is the pitting power $\gamma$. This parameter is important because it determines the level of corrosion according to the material being eroded. However, as we are not dealing with real metal surfaces, this parameter is not known for the image texture analysis. Figure~\ref{fig:parametersusptex} is now studied for $\gamma$ ranging from 0.01 to 0.08. It can be seen that the highest success rate is obtained with $\gamma$ and $\nu$ equal to 0.05 and 1, respectively. For values of $\gamma$ below 0.05 the rate tends to be reduced while for values above 0.05 the rate also tends to decrease. When looking at combined high values of $\gamma$ and $\nu$ the success rates drop sharply. Figure~\ref{fig:iterationsusptex} shows the number of iterations to obtain the highest success rate for each of the parameter combinations. The graph shows that the optimal number of iterations necessary for $\gamma$ equal to 0.05 and $\nu$ equal 1 is relatively low in comparison to the other results.

\begin{figure}[!htbp]
\center
\includegraphics[width=0.8\columnwidth]{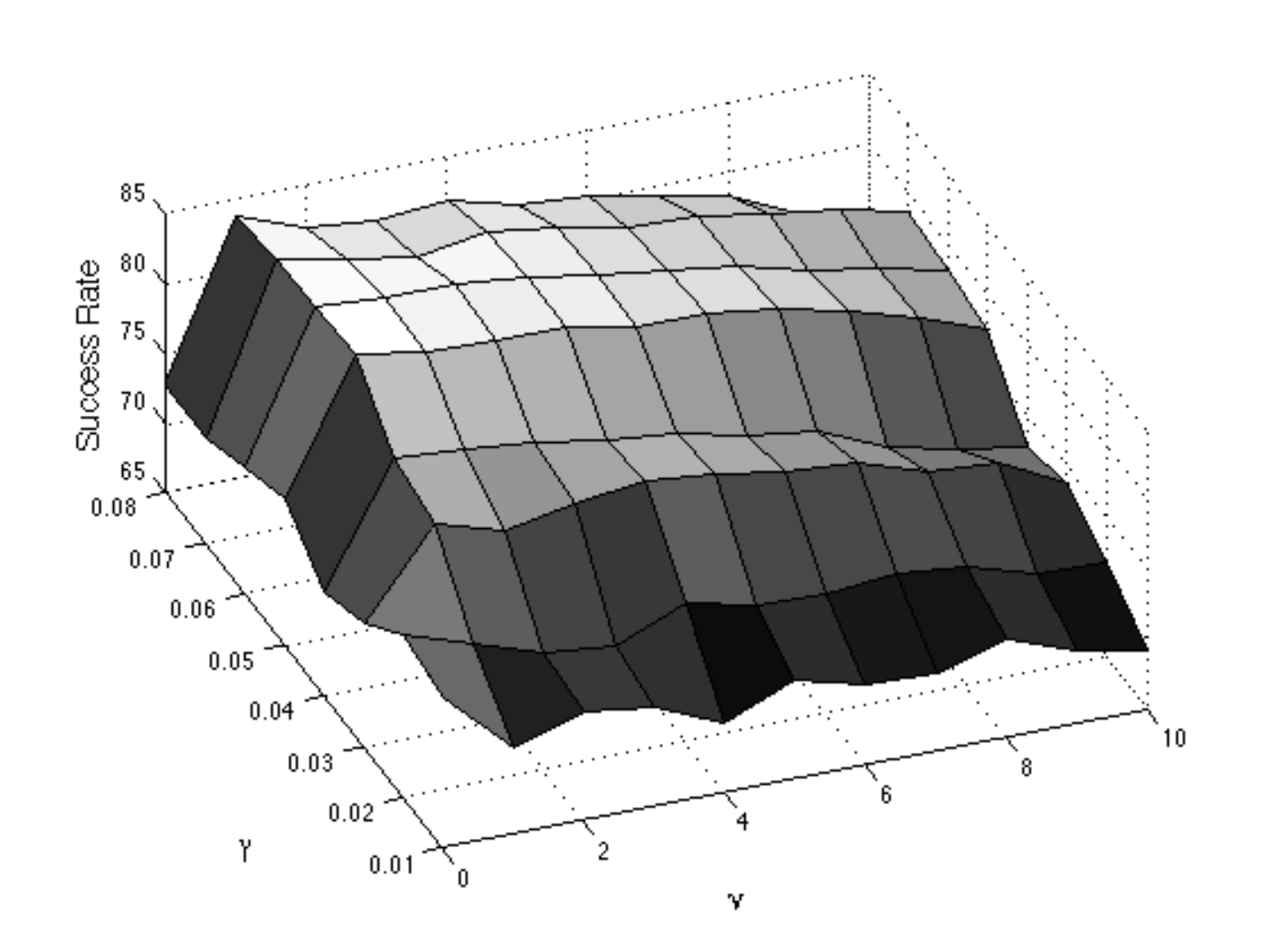}
(a)\\
\includegraphics[width=0.8\columnwidth]{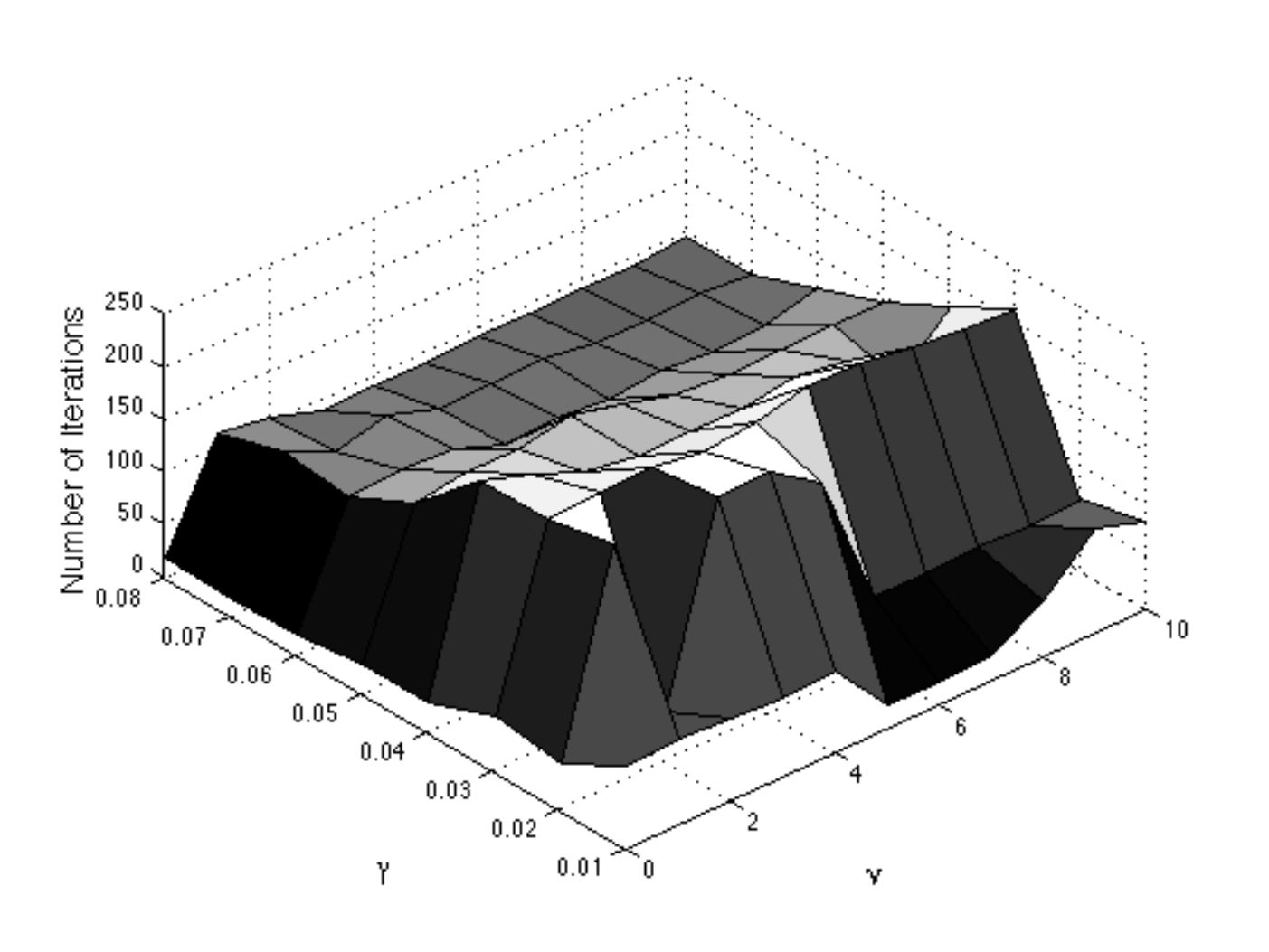}
(b)\\
\caption{ (a) Pitting power and surface roughness analysis for the Usptex database with $\gamma$ from 0.01 to 0.08 and $\nu$ from 0 to 10. (b) Number of iterations for each parameter configuration in (a).} 
\end{figure}

%--------------------------------------------------------------------------------------
\section{Results and Discussions}
\label{sec:resultsDiscussions}

This section reports on the performance of the CITA method for texture analysis. Results for the classification with the proposed method are compared to the traditional texture analysis methods in literature described in Section~\ref{sec:experiments} to evaluate the performance of the method, where the classification is performed using LDA following a stratified 10-fold cross-validation scheme. Three sets of experiments were performed: firstly on the original test databases and afterwards on modified versions of the test databases to test noise and rotation invariance. All tests were performed using the optimized values for $\nu$ and $\gamma$ found in the previous section and the corresponding optimal number of iterations for that specific combination of $\nu$ and $\gamma$, which are shown in Table~\ref{tab:optvalues}.  

\begin{table}[!htbp]
\caption{Optimal parameter values.}
\label{tab:optvalues}
\begin{center}       
\begin{tabular}{lll}
\hline\noalign{\smallskip}
Parameter & Value \\
\noalign{\smallskip}\hline\noalign{\smallskip}
Number of iterations  & 158 \\
$\nu$                        & 1   \\ 
$\gamma$                 & 0.05 \\
\noalign{\smallskip}\hline
\end{tabular}
\end{center}
\end{table}

%--------------------------------------------------------------------------------------
\subsection{Unmodified databases}

Table~\ref{tab:resbrodatzvistex} lists the results for the different texture analysis methods for the unmodified databases. As can be seen, the CITA method achieves an excellent success rate, which outperforms all methods for the Brodatz database and shares the best result for the Vistex database together with the GLDM method. To further test the robustness of the CITA method, firstly, noise is applied to the images and, secondly, a rotation of the images is performed to verify whether the performance of the CITA method persists under these circumstances.

% Results  with other methods
\begin{table}[!htbp]
\scriptsize
\caption{Comparison of the CITA method with traditional texture analysis methods for unmodified databases.}
\label{tab:resbrodatzvistex}
\begin{center}       
\begin{tabular}{lrr}
\hline\noalign{\smallskip}
 & \multicolumn{2}{c}{Success rate (RMSE)}\\
Method & Brodatz &  Vistex \\
\noalign{\smallskip}\hline\noalign{\smallskip}
%stratified 10-fold cross validation
Fourier descriptors   & 94 ($\pm$ 2.2)  & 94 ($\pm$ 1.8)\\
GLCM                       & 94 ($\pm$ 3.1)  & 94 ($\pm$ 3.3)\\
GLDM				      & 98 ($\pm$ 0.9)  & \textbf{97} ($\pm$ 1.7)\\
Gabor filter		      & 92 ($\pm$ 3.5)  & 92 ($\pm$ 1.6)\\
LBPV					      & 88  ($\pm$ 3.2)  & 82 ($\pm$ 3.7)\\
CITA Method 	          & \textbf{99} ($\pm$ 1.5)	& \textbf{97} ($\pm$ 1.6) \\
\noalign{\smallskip}\hline
\end{tabular}
\end{center}
\end{table}

%--------------------------------------------------------------------------------------
\subsection{Noise Invariance}

\begin{figure}[!htbp]
\begin{center}
  \includegraphics[width=\columnwidth]{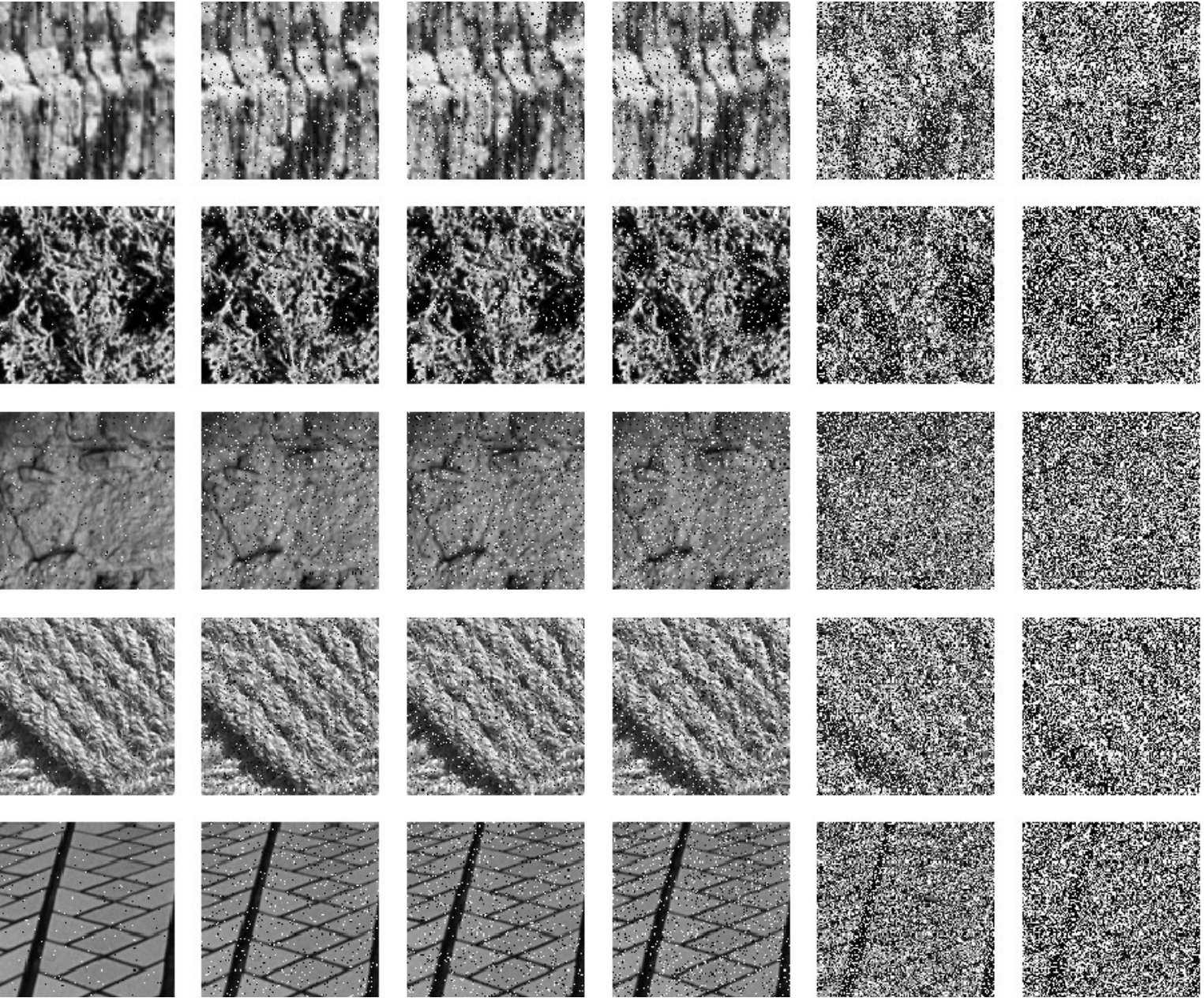}
	\caption{Samples of six databases generated from the Vistex database by adding `Salt \& Pepper'  noise. Each column represents an intensity of noise with $l$ = 0.01, 0.05, 0.07, 0.1, 0.5 and 0.7 from left to the right.}
\label{fig:vistexnoise}  
\end{center}     
\end{figure}

In order to demonstrate the tolerance of the proposed method to noise, experiments were performed on modified versions of the Brodatz and Vistex databases with addition of noise in the form of `Salt \& Pepper' noise. By applying this type of noise to an image, black and white pixels are randomly added to the image matrix with an intensity $l$ which may vary from 0 to 1 and represents the share of the image affected by the noise. The robustness of the CITA method to the addition of noise is demonstrated by performing the texture classification on six modified databases generated from both the Brodatz and Vistex databases. The six different databases were generated in both cases by adding `Salt \& Pepper' noise with intensities $l = 0.01$, 0.05, 0.07, 0.1, 0.5 and 0.7. For all different cases the CITA method is compared with the established methods described in Section~\ref{sec:experiments} in order to get an idea of how the CITA method, in comparison with the other methods, deals with deformation of texture. Figure~\ref{fig:vistexnoise} shows samples of the modified Vistex databases where noise was added to the images and where each column shows examples of an intensity $l$ of noise.

%\begin{figure*}
%\begin{center}
%  \includegraphics[width=0.9\textwidth]{brodatz_noise.pdf}
%	\caption{Samples of six datasets generated from Brodatz by adding ``Salt \& Pepper"  noise. Each column represents a level of noise with densities $r$ = 0.01, 0.05, 0.07, 0.1, 0.5, 0.7 from left to the right.}
%\label{fig:brodatznoise}  
%\end{center}     
%\end{figure*}

%%% comentar resultados

%Table \ref{tab:resbrodatznoise} shows results for six datasets from Brodatz by  adding ``Salt \& Pepper"  noise with different levels. 

The success rates for classifying the perturbated images from the modified Brodatz and Vistex databases using the different methods, are given in Tables~\ref{tab:resbrodatznoise}~and~\ref{tab:resvistexruido}, respectively. These results demonstrate the good performance of our method even with the addition of various intensities of noise. For all databases generated from the Brodatz database, the CITA method has a higher success rate compared to traditional methods in literature. It is important to note that even with increasing noise levels, the CITA method yields high success rates, while for all other methods the success rate declines. For the databases generated from the Vistex database our method gives rise to the second best success rate, preceded by the GLDM method, but still demonstrating its robustness to noise.

%% Result NOISE Brodatz with other methods
\begin{table*}[!htbp]
\scriptsize
\caption{Success rates of texture classification for six databases obtained by adding different intensities $l$ of `Salt \& Pepper' noise to the Brodatz database.}
\label{tab:resbrodatznoise}
\begin{center}       
\begin{tabular}{lrrrrrr}
\hline\noalign{\smallskip}
                                       & \multicolumn{6}{c}{Success rate (RMSE)}\\
Method 					& $l$ = 0.01 &  $l$ = 0.05	& $l$ = 0.07 & $l$ = 0.1 &  $l$ = 0.5	& $l$ = 0.7\\
\noalign{\smallskip}\hline\noalign{\smallskip}
Fourier                 & 93 ($\pm$ 2.5)   &  93  ($\pm$ 3.1)       & 91  ($\pm$ 3.3)      &  90  ($\pm$ 3.3)   & 82  ($\pm$ 4.8)      & 67 ($\pm$ 6.0)\\
GLCM                    & 	94	($\pm$ 3.2)     &  94  ($\pm$ 2.6)       & 95  ($\pm$ 2.6)       &  94  ($\pm$ 2.9)   & 87  ($\pm$ 4.6)      & 76 ($\pm$ 5.9)\\
GLDM                    & 	98    ($\pm$ 1.4)    &  98   ($\pm$ 1.5)      & 98  ($\pm$ 1.4)       &  98  ($\pm$ 1.7)   & 95   ($\pm$ 2.0)      & 91 ($\pm$ 3.4)\\
Gabor filter		    & 	91   ($\pm$ 3.5)    &  90  ($\pm$ 4.3)      & 90   ($\pm$ 4.4)      &  90   ($\pm$ 4.8)  & 82   ($\pm$ 5.5)      & 67 ($\pm$ 3.6)\\
LBPV				 	  & 	88   ($\pm$ 3.0)     &  87  ($\pm$ 3.6)      & 87   ($\pm$ 4.3)      &  87 ($\pm$ 4.5)     & 66   ($\pm$ 5.0)        & 46  ($\pm$ 6.1) \\
CITA method 	  & 	\textbf{99} ($\pm$ 1.6)      &  \textbf{98}   ($\pm$ 1.3)       & \textbf{98}  ($\pm$ 1.3)        & \textbf{98}  ($\pm$ 1.3)    & \textbf{97}  ($\pm$ 2.1)       & \textbf{97}  ($\pm$ 1.9)\\
\noalign{\smallskip}\hline
\end{tabular}
\end{center}
\end{table*}

%% Result NOISE Vistex with other methods
\begin{table*}[!htbp]
\scriptsize
\caption{Success rates of texture classification for six databases obtained by adding different intensities $l$ of `Salt \& Pepper' noise to the Vistex database.}
\label{tab:resvistexruido}
\begin{center}       
\begin{tabular}{lrrrrrr}
\hline\noalign{\smallskip}
                            & \multicolumn{6}{c}{Success rate (RMSE)}\\
Method 					    & $l$ = 0.01 &  $l$ = 0.05	& $l$ = 0.07 & $l$ = 0.1 &  $l$ = 0.5	& $l$ = 0.7\\
\noalign{\smallskip}\hline\noalign{\smallskip}
Fourier                 & 	93	($\pm$ 2.3)     &  91 ($\pm$ 2.3)        & 89 ($\pm$ 1.5)       &  89  ($\pm$ 2.0)    & 67  ($\pm$ 3.5)       & 35 ($\pm$ 5.4)\\
GLCM                    & 	94	($\pm$ 2.8)     &  95 ($\pm$ 1.6)        & 95 ($\pm$ 2.7)        &  94 ($\pm$ 1.7)    & 84  ($\pm$ 3.6)      & 66 ($\pm$ 4.1)\\
GLDM                    & 	\textbf{98}  ($\pm$ 1.1)     &  \textbf{97}  ($\pm$ 1.4)       & \textbf{97}  ($\pm$ 1.5)       &  \textbf{97}     ($\pm$ 1.3) & \textbf{94}  ($\pm$ 1.3)       & \textbf{86} ($\pm$ 2.9)\\
Gabor filter		    & 	91 ($\pm$ 2.1)       &  89  ($\pm$ 1.7)       & 89 ($\pm$ 3.0)      &  86  ($\pm$ 3.1)   & 56  ($\pm$ 5.5)       & 34 ($\pm$ 3.6)\\
LBPV				 	  & 	83 ($\pm$ 3.6)      &  83   ($\pm$ 2.2)      & 82  ($\pm$ 3.8)      &  81  ($\pm$ 3.9)   & 54   ($\pm$ 2.9)      & 39 ($\pm$ 7.3)\\
CITA method 	  & 	96 ($\pm$ 1.7)      &   94  ($\pm$ 1.3)      & 94  ($\pm$ 1.5)      &  94  ($\pm$ 1.4)   & \textbf{94} ($\pm$ 4.4)         & 81 ($\pm$ 4.2)\\
\noalign{\smallskip}\hline
\end{tabular}
\end{center}
\end{table*}

%% TRAINING WITH ORIGINAL SAMPLES AND TEST WITH NOISE
\npar
The results in Tables~\ref{tab:resbrodatznoise}~and~\ref{tab:resvistexruido} are obtained with addition of noise to both the training as well as the test data. Further, experiments using non-perturbated texture images for training and images with addition of noise for testing were performed. The results are shown in Tables~\ref{tab:resbrodatztestnoise}~and~\ref{tab:resvistextestnoise}. The success rate of the CITA method is comparable to the other methods analyzed in both cases. It is never the worst method seen over the different intensities and two databases, but neither is it clearly the best method.

%% Result NOISE Brodatz with other methods
\begin{table*}[!htbp]
\scriptsize
\caption{Success rates of classification of the texture images of the Brodatz database, with training data without addition of noise and test data with addition of noise at different intensities $l$ of `Salt \& Pepper' noise.}
\label{tab:resbrodatztestnoise}
\begin{center}       
\begin{tabular}{lrrrrrr}
\hline\noalign{\smallskip}
                                       & \multicolumn{6}{c}{Success rate (RMSE)}\\
Method 					& $l$ = 0.01 &  $l$ = 0.05	& $l$ = 0.07 & $l$ = 0.1 &  $l$ = 0.5	& $l$ = 0.7\\
\noalign{\smallskip}\hline\noalign{\smallskip}
Fourier                 & 	90 ($\pm$ 2.4)	   &    \textbf{62}  ($\pm$ 3.4)   &   \textbf{50} ($\pm$ 3.8)  &  \textbf{36} ($\pm$ 2.4)  & 4 ($\pm$ 0.9) & 1 ($\pm$ 1.2)\\
GLCM                   & 	49  ($\pm$ 2.0)    &      6 ($\pm$ 0.9)   &       4 ($\pm$ 0.4)   &   3 ($\pm$ 0.6)  &  1 ($\pm$ 0.0) & 1 ($\pm$ 0.0)\\
GLDM                   & 	84  ($\pm$ 5.0)    &    42 ($\pm$ 3.0)   &     16 ($\pm$ 1.5)  &   9 ($\pm$ 0.6)  &  1 ($\pm$ 0.0) & 1 ($\pm$ 0.0)\\
Gabor filter		   & 	73  ($\pm$ 4.7)    &    57 ($\pm$ 1.8)  &      32 ($\pm$ 1.5)  &  26 ($\pm$ 2.1) &  \textbf{7} ($\pm$ 1.1) &  4 ($\pm$ 0.8)\\
LBPV				 	   & 	57  ($\pm$ 3.2)    &    14  ($\pm$ 1.2)  &     10 ($\pm$ 1.3)   &    9 ($\pm$ 0.7) &  1 ($\pm$ 0.0) &  1 ($\pm$ 0.0)\\
CITA method 	   & 	\textbf{97}  ($\pm$ 2.6)     &   37 ($\pm$ 3.2)    &    20 ($\pm$ 1.6)   &  11 ($\pm$ 1.3)  & 6 ($\pm$ 1.8) & \textbf{4} ($\pm$ 0.7)\\
\noalign{\smallskip}\hline
\end{tabular}
\end{center}
\end{table*}

%% Result NOISE Vistex with other methods
\begin{table*}[!htbp]
\scriptsize
\caption{Success rates of classification of the texture images of the Vistex database, with training data without addition of noise and test data with addition of noise at different intensities $l$ of `Salt \& Pepper' noise.}
\label{tab:resvistextestnoise}
\begin{center}       
\begin{tabular}{lrrrrrr}
\hline\noalign{\smallskip}
                            & \multicolumn{6}{c}{Success rate (RMSE)}\\
Method 					    & $l$ = 0.01 &  $l$ = 0.05	& $l$ = 0.07 & $l$ = 0.1 &  $l$ = 0.5	& $l$ = 0.7\\
\noalign{\smallskip}\hline\noalign{\smallskip}
Fourier                 & 	58 ($\pm$ 4.1)	   &   10  ($\pm$ 2.0)    &   8 ($\pm$ 2.0) &    6 ($\pm$ 2.0) &   3 ($\pm$ 1.1) & 2.1 ($\pm$ 1.0) \\
GLCM                   & 	64 ($\pm$ 3.5)     &    \textbf{33}  ($\pm$ 1.4)    &    5 ($\pm$ 0.5) &    2 ($\pm$ 0.6)  &  2 ($\pm$ 0.6)  &  2 ($\pm$ 0.6)\\
GLDM                   & 	\textbf{90} ($\pm$ 2.5)     &    13 ($\pm$ 1.4)  &    13 ($\pm$ 1.1) &   9 ($\pm$ 1.2) &   2 ($\pm$ 0.5) &   2.3 ($\pm$ 1.0)\\
Gabor filter		   & 	58 ($\pm$ 2.6)    &    14 ($\pm$ 3.2)  &     10 ($\pm$ 2.1) &     5. ($\pm$ 1.5) &   2 ($\pm$ 0.5) &  2 ($\pm$ 0.6)\\
LBPV				 	   & 	52  ($\pm$ 2.6)    &    14 ($\pm$ 2.5)  &     7 ($\pm$ 1.5) &     5.6 ($\pm$ 1.5) &  2 ($\pm$ 0.6) &  2 ($\pm$ 0.6)\\
CITA method 	   & 	83  ($\pm$ 3.2)    &    23 ($\pm$ 4.1)  &    \textbf{17} ($\pm$ 2.5) &    \textbf{ 11} ($\pm$ 1.6) &    \textbf{6} ($\pm$ 3.1) &   \textbf{5} ($\pm$ 2.8)\\
\noalign{\smallskip}\hline
\end{tabular}
\end{center}
\end{table*}

%--------------------------------------------------------------------------------------
\subsection{Rotation Invariance}

\begin{figure}[!htbp]
\begin{center}
  \includegraphics[width=\columnwidth]{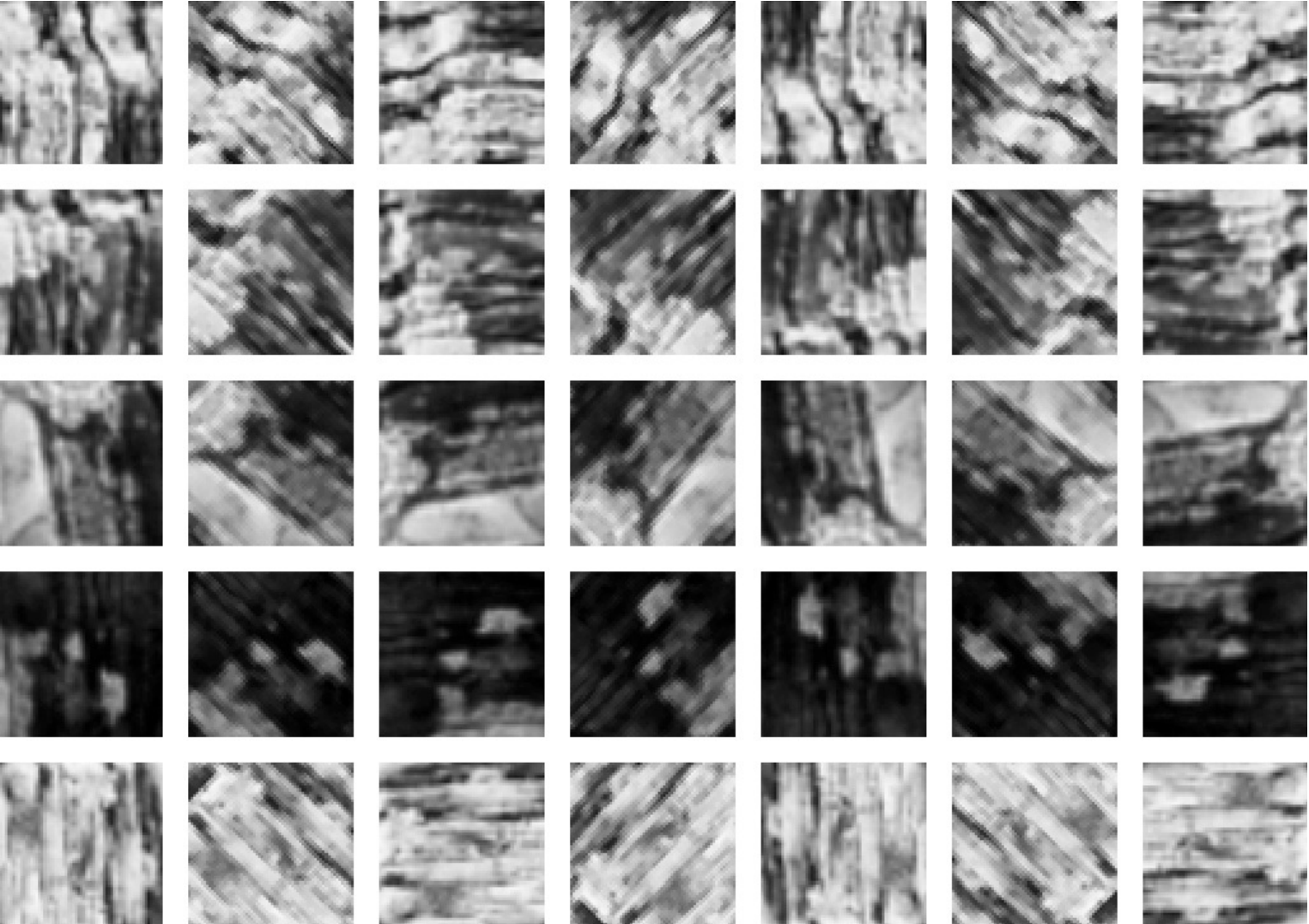}
	\caption{Samples of rotated images from the Brodatz database. Each column corresponds to a different rotation angle. From left to right: 0$^{\circ}$, 45$^{\circ}$, 90$^{\circ}$, 135$^{\circ}$, 180$^{\circ}$, 225$^{\circ}$ and 270$^{\circ}$.}
\label{fig:brodatzrot}  
\end{center}     
\end{figure}

The proposed CITA method is intrinsically rotation invariant, so good results are expected when test are performed with modified databases with rotated images. To demonstrate the rotation invariance of the CITA method, additional versions of both the Brodatz and Vistex databases are created. Each image from the databases is rotated with the following angles: 0$^{\circ}$, 45$^{\circ}$, 90$^{\circ}$, 135$^{\circ}$, 180$^{\circ}$, 225$^{\circ}$ and 270$^{\circ}$ and in this way, seven images are obtained from each original database image. Therefore, the new database with rotated Brodatz images has 70 images per class with 111 classes in total and the new database with rotated Vistex images has 112 images per class with 54 classes in total. Figure~\ref{fig:brodatzrot} shows for some texture images from the Brodatz database the seven rotated images obtained under the different rotation angles, with all images on the same row originating from the same original image. 

%\begin{figure*}
%\begin{center}
%  \includegraphics[width=0.9\textwidth]{vistex_rot.pdf}
%	\caption{Samples (200 $\times$ 200 pixels size) of rotated images from Vistex. Each column represents the 
%correspondingly angles from left to the right: 0$^{\circ}$, 45$^{\circ}$, 90$^{\circ}$, 135$^{\circ}$, 180$^{\circ}$, 225$^{\circ}$ and 270$^{\circ}$.}
%\label{fig:vistexrot}  
%\end{center}     
%\end{figure*}

\npar
Table~\ref{tab:resrotated} shows the success rates for the classification of the texture images of the Brodatz and Vistex databases with rotated images. In this case, the success rates are obtained with LDA following a stratified 10-fold cross-validation scheme for the complete rotated Brodatz and Vistex databases, where each database consists of all rotated texture images of all the original images. The success rates achieved with the CITA method are better than the success rates obtained with any of the other methods and are comparable to the results obtained on the unmodified databases. These experimental results indicate that our method has a good generalization ability. Hence, the method described here has proven to be performant also for rotated texture classification.

%% Result Rotated Brodatz and Vistex with other methods
\begin{table}[!htbp]
\scriptsize
\caption{Comparison of the CITA method with traditional texture analysis methods for the databases of rotated images.}
\label{tab:resrotated}
\begin{center}       
\begin{tabular}{lrr}
\hline\noalign{\smallskip}
             & \multicolumn{2}{c}{Success rate (RMSE)}\\
Method & Rotated Brodatz &  Rotated Vistex \\
\noalign{\smallskip}\hline\noalign{\smallskip}
Fourier descriptors         & 83 ($\pm$ 1.0) &  77 ($\pm$ 1.3)\\
GLCM 	                         & 71 ($\pm$ 0.5) &  74 ($\pm$ 1.2)\\
GLDM					         & 84 ($\pm$ 0.5) & 83 ($\pm$ 0.7)\\
Gabor filter		             & 78 ($\pm$ 0.9) & 71 ($\pm$ 1.3)\\
LBPV				 	             & 63 ($\pm$ 1.0) & 62 ($\pm$ 1.6)\\
CITA method 	 	        &  \textbf{98} ($\pm$ 0.4) &  \textbf{97} ($\pm$ 0.3)\\
\noalign{\smallskip}\hline
\end{tabular}
\end{center}
\end{table}

%--------------------------------------------------------------------------------------
\section{Conclusions}
\label{sec:conclusions}

In this paper, a new method for texture analysis and classification was described. Combining concepts from corrosion engineering, cellular automata and pattern recognition a texture descriptor was generated, able to characterize an image according to the iterations of a CA-based model drawing inspiration from the pitting corrosion phenomenon. The developed CITA method was used to classify the texture images of two well-known databases: Brodatz and Vistex. The method was applied to images of the original databases and the robustness of the method under addition of noise and rotation was investigated. For this purpose, several new databases were created, starting from the original databases. Six new databases were obtained by adding `Salt \& Pepper' noise with different intensities to each of the images of the test databases and another new database was obtained by rotating the images of the databases under seven angles. In all cases, the CITA method obtained good results compared to the methods from literature, showing a good generalization ability and proving to be performant for texture classification. 

\npar
The results presented in this paper demonstrate the potential of the CITA method. Therefore, future work should focus on further refining the method as well as expanding it so that it is applicable for more types of texture images. This can be done by integrating measures of corrosion frequency via a histogram of eroded pixels and measuring the velocity of corrosion by calculating the difference between initial and final values divided by the number of iterations. Further, the CITA method has to be expanded so that it can deal with RGB color images as well as with dynamic textures, i.e. sequences of images that together form a texture.

%--------------------------------------------------------------------------------------
\section*{Acknowledgments}

N\'ubia Rosa da Silva acknowledges support from FAPESP (The State of S\~ao Paulo Research Foundation). Pieter Van der Wee\"{e}n was sponsored by the Fund for Scientific Research in Flanders (FWO). Odemir Martinez Bruno gratefully acknowledges the financial support of CNPq (National Council for Scientific and Technological Development, Brazil) (Grant Nos. 308449/2010-0 and 473893/2010-0) and FAPESP (Grant No. 2011/01523-1).

%\bibliography{texture}

\end{document}